\documentclass[%
 reprint,
 amsmath,amssymb,
 showkeys,
]{revtex4-2}

\usepackage{graphicx,epsfig,amsmath,color}
\usepackage{dcolumn}
\usepackage{bm}

\usepackage{hyperref}
\usepackage{url}
\usepackage{graphicx,epsfig,amsmath,color}
\newcommand{\beq}{\begin{equation}}
\newcommand{\eeq}{\end{equation}}
\newcommand{\beqn}{\begin{eqnarray}}
\newcommand{\eeqn}{\end{eqnarray}}
\def\bmath#1{\mbox{\boldmath$#1$}}
\DeclareMathOperator*{\argmin}{arg\,min}
\DeclareMathOperator*{\argmax}{arg\,max}

\usepackage{algorithm}
\usepackage{algpseudocode}

\def\mnras{Monthly Notices of the Royal Astronomical Society} 
\def\aap{Astronomy and Astrophysics}
\def\prl{Physical Review Letters}
\def\apj{Astrophysical Journal}

\begin{document}

\preprint{APS/XXX}

\title{Recovering Weak Signals with Normalizing Flows}

\author{Sarod Yatawatta}
 \email{yatawatta@astron.nl}
\affiliation{%
  ASTRON, Netherlands Institute for Radio Astronomy,\\ Oude Hoogeveensedijk 4, 7991 PD, Dwingeloo, The Netherlands.\\
}%

\date{\today}

\begin{abstract}
  In many scientific disciplines, weak signals of interest are obscured by dominant nuisance signals that are several orders of magnitude stronger. Recovering these weak signals requires subtracting the dominant ones; however, this calibration process inherently distorts or partially suppresses the underlying signal of interest. To address this problem, we propose the use of normalizing flow models to reconstruct calibration-affected weak signals. By leveraging the statistical invariance of the target signals and assuming minimal initial suppression, our framework effectively recovers the lost signal components. We provide a comprehensive theoretical overview of this normalizing flow-based recovery method and demonstrate its efficacy using simulated data.
\end{abstract}

\maketitle


\section{Introduction\label{sec:intro}}
Many disciplines in physics rely on the accumulation of observational data and the recovery of weak, hidden signals in that data. A case in point are the signals that originate from the Universe that feed several research areas in cosmology. Such signals have statistical stability as an advantage but they are overwhelmed by nuisance signals that are of many orders of magnitudes higher in power level (e.g., Galactic and extra-Galactic foregrounds). 

\begin{figure}[ht]
  \begin{minipage}{0.99\linewidth}
    \begin{center}
  \epsfig{figure=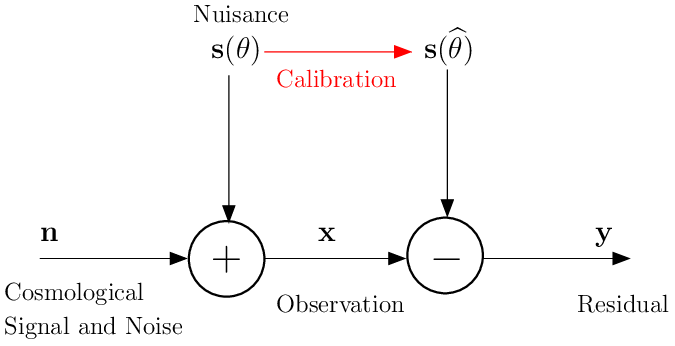,width=8.0cm}\\
    \end{center}
  \end{minipage}
  \caption{A schematic of the signal flow. The nuisance signals are subtracted from the observed data $\bm x$ by calibration to get the residual $\bm y$.\label{fig:block}}
\end{figure}

The estimation and the removal of nuisance signals from the observed data are performed by many specialized techniques unique to each scientific discipline but in this paper, we broadly consider them to be a form of calibration. In calibration, a parameterized model for the nuisance signals are constructed and subtracted from the data (see Fig. \ref{fig:block}). The desired weak signals are assumed to remain intact in the residual. Due to the statistical stability of the weak, hidden signals, it is generally assumed that the residual data can be accumulated without any limit to reach the desired dynamic range for their detection.

The statistical performance of calibration (in radio interferometry) has been studied both theoretically \citep[e.g.,][]{ST2019,Millad2018} and empirically \citep[e.g.,][]{Patil2016,EW2017,mevius2021}. Its effect on the residual and on the weak signals hidden in the residual is well known. However, no method exists that can recover the original weak signals hidden in the residual. This work intends to improve on that.

In this paper, we propose the use of normalizing flows \citep{Papamakarios} to recover the undistorted weak signals in the residual after calibration. 
Normalizing flows have been used in many data processing applications such as cosmological parameter estimation and model construction
\citep[e.g.,][]{Langendorff2023,Baso2022,Dome2024,Pund2024,Sun2025} and simulations \citep[e.g.,][]{rouhiainen2021normalizing,mebratu2026wavelet,Hassan2022,Moot2025}. There are many more potential application areas \citep{Denzel2026,Acharya2026} and our work covers one such application. The statistical relationship between the input data and the output residual can be described by using the influence function \citep{Hampel86,cook1982residuals,Koh17,ST2019}. This relationship enables the closed form calculation of the Jacobian of the mapping between the input and the output residuals. We incorporate this into a normalizing flow model to train the base distribution which enables the determination of the statistics (mainly first order) of the hidden weak signals.

Existing work based on flow models mostly learn the Jacobian (or its inverse) as part of the training process. In contrast, we work with a flow model where the Jacobian of the mapping is externally provided and hence we are not constrained to a bijective mapping. Calibration is a computationally demanding task in data processing and this is performed by specialized and highly optimized software to cope with large data volumes. As a bi-product of calibration, the Jacobian of the mapping from data to residual is also produced. In contrast, common applications of normalizing flows  \citep[e.g., contractive residual flows][]{chen2019residual} need to learn it themselves. Parameter estimation in (ill-constrained) linear models has some similarity to this work. Prior distributions are trained using flow models in \citep{wei2022deep} while stochastic differential equations for structured noise are combined with parameter estimation in \citep{Stevens2025}. Contrastive flows for signal and background separation (in event classification) are studied in \citep{elsharkawy2025} where the background signal is considered a nuisance parameter. A separate flow model for nuisance parameters is trained in \citep{melnychuk2023} for interventional (causal) density estimation. Similarly, in \citep{Valsecchi2026}, factorized flow models are used to directly estimate the distribution of interest as well as the nuisance parameters. Note that in our method, the nuisance parameters are handled by calibration, without having to train a flow model for computational efficiency. Normalizing flows that flow in both directions (from data to noise and from noise to data) are considered in \citep{abdelhamed2019} that uses learnt approximate inverse mappings for heteroscedastic noise modeling in image processing. Flow models with non-invertible Jacobians are considered in \citep{flouris2023canonical}  where the flow is embedding the data onto a low dimensional manifold. 

In this paper, we propose a method that uses a normalizing flow for density estimation where the Jacobian of the forward mapping is explicitly provided. Existing work use normalizing flows for profiling systematic uncertainties in simulation-based inference or act as explicit priors for inverse problems. In this work, we focus on the structural degradation of the signal of interest downstream by calibration. We consider the signal in the traditional sense to be a nuisance (Fig. \ref{fig:block}) and the true signal of interest to be hidden in the noise. We do not have the freedom to select the nuisance model (that depends on the physics of the observational setup). The nuisance model is generally considered to be non-linear and non-invertible. A major assumption of our method is that the signal hidden in the noise is weak compared to the nuisance signal and the noise is independent of the nuisance signal. Moreover, we assume statistically stable noise compared to the nuisance signal that may have variations temporally, spatially and spectrally (for example, we can consider the noise to be of cosmological origin while the nuisance to be of Galactic and instrumental origin). Under these assumptions (that are reasonable for experiments measuring signals of cosmological origin), we illustrate a method to recover the weak signals hidden in the noise in this paper. The remainder of this paper is organized is as follows: in section \ref{sec:data_model}, we provide a theoretical overview of calibration and the statistical relationship between the input and the output residual. In section \ref{sec:norm}, we provide an overview of normalizing flows and the setup we use to train the flow model. Next in section \ref{sec:results}, we provide results using a linear regression example as well as using a non-linear example (radio interferometric calibration) before drawing our conclusions in section \ref{sec:conc}.
 
{\em Notation}: Lowercase bold letters refer to column vectors (e.g., ${\bm y}$). Uppercase bold letters refer to matrices (e.g., ${\bm C}$). Unless otherwise stated, all variables are real numbers. The set of complex numbers is given as ${\mathbb C}$ and the set of real numbers is given as  ${\mathbb R}$. The matrix inverse, pseudo-inverse, transpose, and Hermitian transpose are referred to as $(\cdot)^{-1}$, $(\cdot)^{\dagger}$, $(\cdot)^{T}$, and $(\cdot)^{H}$, respectively. The matrix Kronecker product is given by $\otimes$. The identity matrix is given by ${\bf I}$ (size depends on the context). The Frobenius norm (for a matrix) or the L-$2$ norm (for a vector) is given by $\|\cdot \|$ and the L-$1$ norm is given by $\|\cdot\|_1$. The vectorization of a matrix is represented by $\mathrm{vec}(\cdot)$.

\section{Data Model\label{sec:data_model}}
We consider the data flow illustrated in Fig. \ref{fig:block}.
The observed data are denoted as $\bm{ x}$ ($\in \mathbb{R}^D$), which consists of a parametric model ${\bm s}(\bm {\theta})$ (${\bm s}(\cdot)$ is a mapping from $\mathbb{R}^{M}$ to $\mathbb{R}^{D}$) and additive noise $\bm n$ ($\in \mathbb{R}^D$). The unknown parameters $\bm \theta$ ($\in \mathbb{R}^M$) parameterize the nuisance signal for example by modeling the foreground signals and instrumental systematics. The observed data is given by (\ref{eq:obs})
\beq \label{eq:obs}
\bm{x} = {\bm s}(\bm{\theta}) + \bm{n}.
\eeq

During calibration, the systematics model parameters $\bm {\theta}$ are estimated, for example by using maximum likelihood estimation or any similar technique to get $\widehat{\bm{\theta}}$
\beq \label{eq:mle}
\widehat{\bm{\theta}}=\underset{\bm{\theta}}{\argmin} f(\bm {x};\bm{\theta})
\eeq
where $f(\bm{ x};\bm{\theta})$ is the negative likelihood or similar function that is minimized to estimate $\bm{\theta}$. Note that the noise $\bm{n}$ is assumed to have a known distribution (for example zero mean white Gaussian noise) during solving (\ref{eq:mle}). This is only possible because the weak signal hidden in ${\bm n}$ is hardly noticeable compared to the power of the nuisance signal $s(\bm{\theta})$, i.e.,  $\| {\bm s}(\bm{\theta})  \| \gg E\{ {\bm n} \}$. Due to the same reason, in order to extract the weak signal of scientific interest in the noise, we need to find the residual $\bm{y}$,
\beq \label{eq:res}
\bm{ y} = \bm{ x} - {\bm s}(\widehat{\bm{\theta}}).
\eeq

The hope is that the statistics of the residual $\bm {y} \sim p_{Y}(\bm {y})$ are similar to the statistics of the noise ${\bm n} \sim p(\bm{n})$ and therefore, we are ably to study the weak signals of interest. However, in reality, this is not the case. Consider the mapping from the data $\bm {x}$ to the residual $\bm {y}$
\beq
\bm{y} = \bm{T}(\bm{x})
\eeq
where $\bm{T}(\cdot)$ is a mapping from from $\mathbb{R}^{D}$ to $\mathbb{R}^{D}$ and if  $\bm{y} \sim p_{Y}(\bm{y})$ and  $\bm{x} \sim p_{X}(\bm{x})$, we have
\beq \label{eq:pxy}
p_{X}(\bm{x}) = |\bm J|\ \ p_{Y}(\bm{y})
\eeq
where $\bm J$ is the Jacobian of the mapping $\bm T(\cdot)$, i.e., $\bm J=\frac{\partial \bm T({\bm{x}})}{\partial {\bm{x}}^T}$ ($\in \mathbb{R}^{D\times D}$) and $|\bm J|$ is the absolute value of its determinant.

From (\ref{eq:obs}), we also see that
\beq \label{eq:pnx}
p(\bm{n})=p_X(\bm{x})
\eeq
because the noise is independent of $\bm s(\bm{\theta})$ and $\frac{\partial {\bm s}(\bm{\theta})}{\partial \bm{n}^T}=\bm{0}$.

From (\ref{eq:pxy}) and (\ref{eq:pnx}) we see that it is possible to recover $p(\bm{n})$ and extract the weak signals hidden therein (we focus exclusively on $E\{{\bf n}\}$).

Before proceeding further, we state the assumptions:
\begin{itemize}
  \item The nuisance signal is dominant compared to the weak signal hidden in noise, i.e., $\| {\bm s}(\bm{\theta})  \| \gg E\{ {\bm n} \}$.
  \item The residual $\bf y$ is close to the noise $\bf n$, i.e., $||{\bm s}(\bm{\theta})- {\bm s}(\widehat{\bm{\theta}})|| \approx 0$.
  \item At the solution of (\ref{eq:mle}), a local minimum of the cost function exists, i.e., $\frac{\partial f(\bm{x};\bm{\theta})}{\partial\bm{\theta}^T}={\bm 0}$
   \item We have the ability accumulate multiple observations where in each observation the parameters $\bm \theta$ change, creating diversity. For example, in radio interferometry, the rotation of the Earth as well as the variation of systematic errors over time and frequency creates this condition.
  \item The base distribution $p(\bm{n})$ remains statistically stable in contrast to ${\bm s}(\bm{\theta})$ for multiple observations. In radio interferometry, spatial variation is compensated by a phase shift in Fourier space. This (complex) phase shifting matrix is diagonal with complex exponentials in the diagonal, making its determinant unity. Hence for this particular case, any phase shifting does not affect the form of $p(\bm{n})$.
\end{itemize}

Under the aforementioned assumptions, it has been shown by many \citep[e.g.,][]{SAM2018} that
\beq \label{eq:jacobian}
{\bm J}({\bm x},{\bm \theta}) =\bm I + \frac{\partial {\bm s}(\bm \theta)}{\partial\bm \theta^T} \left(\frac{\partial^2 f(\bm{x}; \bm{\theta})}{\partial\bm \theta \partial\bm \theta^T}\right)^{\dagger} \frac{\partial^2 f(\bm{x}; \bm{\theta})}{\partial \bm \theta \partial \bm{x}^T }
\eeq
evaluated at $\widehat{\bm \theta}$ or at the local minimum of $f(\bm{x};\bm{\theta})$. The proof of (\ref{eq:jacobian}) is given in appendix \ref{app:influence}. Note also that (\ref{eq:jacobian}) is in fact the influence function \citep{Hampel86,Koh17}. We see that the residual statistics $p_{Y}(\bm{y})$ are not the statistics of the weak signal we seek. Furthermore, the Jacobian ${\bm J}({\bm x},\widehat{\bm \theta})$ ($\in \mathbb{R}^{D\times D}$) is not necessarily full rank of $D$ (can lose rank at most by the number of degrees of freedom used by $\bm \theta$, i.e., $M$) and hence not necessarily invertible.

The conundrum we face is extracting the statistics of $p_{X}(\bm{x})$ or $p_{N}(\bm{n})$ from the residual $\bm{y}$ and the a-priori Jacobian ${\bm J}({\bm x},\widehat{\bm \theta})$. In section \ref{sec:norm}, we will explore the use of normalizing flows for this purpose. 

\section{Normalizing flows\label{sec:norm}}
Normalizing flows provide a mechanism to train a deep neural network (DNN) to model probability density functions undergoing a transform \citep{Dinh2014,Jimenez2015,Papamakarios}. In our case (see Fig. \ref{fig:block}), we consider the noise density $p(\bm n)$ undergoing a transformation due to calibration to reach the density of the residual, $p_{Y}(\bm{y})$.  
We consider the base distribution $p_{X}(\bm{x})$ or $p(\bm{n}; {\bm{\psi}})$ to be parameterized by ${\bmath \psi}$.
From density transform, we have the log likelihood of $\bm{y}$
\beqn \label{LL}
\log p_{Y}(\bm{y}) =\log p_{X}(\bm{x}) - \log |{\bm J}({\bm x},\widehat{\bm \theta})| \\\nonumber
\propto \log p(\bm{n} ; {\bmath \psi}) - \log |{\bm J}({\bm x},\widehat{\bm \theta})|.
\eeqn

The eigenvalues of the Jacobian ${\bm J}({\bm x},\widehat{\bm \theta})$ are used to efficiently evaluate $\log |{\bm J}({\bm x},\widehat{\bm \theta})|$, i.e.,  the logarithm of the absolute value of the determinant $|{\bm J}({\bm x},\widehat{\bm \theta})|$.
Direct maximization of (\ref{LL}) for likelihood maximization is ill constrained, mainly because ${\bm T}(\bm{x})$ in (\ref{eq:pxy}) is not necessarily invertible. Normalizing flows with non-invertible Jacobians have been used with additional constraints for improved stability \cite{flouris2023canonical}. Following the same procedure, we introduce an additional constraint such that the residual $\bm y$ is close to $\bm{n}$; in other words the correlation $\bm{n}^T \bm{y}$ is high. With this additional constraint, we formulate the loss function to maximize as
\beqn \label{loss1}
\bm{\psi} = \underset{\bm \psi}{\argmax}\ \ \log p(\bm{n} ; {\bm{\psi}}) - \log |{\bm J}({\bm x},\widehat{\bm \theta})|\\\nonumber
\mathrm{subject\ to}\ \  \bm{n}^T \bm{y} > \gamma
\eeqn
where $\gamma \in \mathbb{R}^{+}$ is the lower bound for the expected correlation between ${\bm n}$ and ${\bm y}$. We use an inequality constraint in (\ref{loss1}) instead of a penalty term or an equality constraint because we do not have a measure of the exact correlation between ${\bf n}$ and ${\bm y}$. With inequality constraints, we use the augmented Lagrangian method \citep{Giesen} where we introduce
\beq \label{gfun}
g(\bm{n},\bm{y}) \buildrel \triangle \over =\left(\left[\gamma -  \frac{\bm{n}^T \bm{y}}{\overline{\| \bm{n} \|}\ \overline{\| \bm{y} \|}}\right]_{+}\right)^2
\eeq
making $g(\cdot,\cdot)$ active only when positive to satisfy the constraint, where $[\cdot]_{+}$ is the ReLU() operation. Note that we find the normalized inner product between $\bm{n}$ and $\bm{y}$ where $\overline{\| \bm{n} \|}$ and $\overline{\| \bm{y} \|}$ denote the norms of $\bm{n}$ and $\bm{y}$ averaged over the minibatch. In this manner, we overcome the need to adjust $\gamma$ too much depending on the problem.

We consider stochastic training where we have a minibatch $\mathcal{B}$ to sample data. With (\ref{gfun}), we have the augmented Lagrangian for minibatch $\mathcal{B}$ 
\beqn \label{total_loss}
\lefteqn{L({\bmath \psi})=}\\\nonumber
&&E_{\bm{y},\bm{J}({\bm x},\widehat{\bm \theta})\sim \mathcal{B}, \bm{n}\sim p(\bm{n}; {\bmath \psi})}\left[- \log p(\bm{n} ; {\bmath \psi})+ \log |\bm J({\bm x},\widehat{\bm \theta})|\right. \\\nonumber
&&\left. + \frac{\rho}{2} g(\bm{n},\bm{y})^2 +  z\ g(\bm{n},\bm{y}) \right]
\eeqn
where $\rho \in \mathbb{R}^+$ is the regularization factor and $z \in \mathbb{R}$ is the Lagrange multiplier. After several updates of $\bmath \psi$, we update the Lagrange multiplier $z$ as
\beq \label{zupdate}
z \leftarrow  z + \rho\ g(\bm{n},\bm{y}).
\eeq

In certain situations, pre-training the model with the loss
\beq \label{preloss}
l({\bmath \psi}) =  E_{\bm{y} \sim \mathcal{B}, \bm{n}\sim p(\bm{n}; {\bmath \psi})}\left[ \| \bm{n} - \bm{y} \|^2 \right]
\eeq
will be helpful before training the normalizing flow.

In Algorithm \ref{algN}, we present the pseudocode for training the normalizing flow model. More practical details such as using warmup and learning rate schedulers are presented in Appendix \ref{app:training}. In section \ref{sec:results}, we present results using two examples.
\begin{figure}
\begin{algorithm}[H]
  \caption{Train normalizing flow}\label{algN}
\begin{algorithmic}[1]
  \Require Epochs $e$, batch size $b$, cadence $C$, $\gamma$,$\rho$ 
  \State Initialize $z \leftarrow 0$, $\bmath \psi$
  \If{pre training is enabled}
    \For{Epoch in $e$}
      \For{Batch $\mathcal{B}$}
      \State Update ${\bmath \psi}$ by minimizing (\ref{preloss})
      \EndFor
    \EndFor
  \EndIf
  \State $n_i\leftarrow 1$
  \For{Epoch in $e$}
      \For{Batch $\mathcal{B}$}
      \State Update ${\bmath \psi}$ by minimizing (\ref{total_loss})
      \If{$n_i$ is a multiple of $C$}
      \State Update $z$ using (\ref{zupdate})
      \EndIf
      \State $n_i\leftarrow n_i+ 1$
      \EndFor
  \EndFor
\State Return solution $\bmath \psi$
\end{algorithmic}
\end{algorithm}
\end{figure}

\section{Simulation Results\label{sec:results}}
We provide two examples where we seek a weak signal hidden in the noise. We first provide a simple linear example and next a complicated example based on radio interferometry. Note that the nuisance models ${\bm s}(\bm{\theta})$ in (\ref{eq:obs}) are vastly different in the two examples but we can apply the same method as summarized in Algorithm \ref{algN} for both.
\subsection{Linear model\label{sec:linear}}
We consider the nuisance model to be linear, i.e.,
\beq \label{ex_data}
{\bm x}={\bm A}{\bmath \theta} + {\bm n}
\eeq
where ${\bm A}$ ($\in \mathbb{R}^{D\times M}$) is the design matrix that is known for each observation of ${\bm x}$ ($\in \mathbb{R}^D$) and ${\bm \theta}$ ($\in \mathbb{R}^{M}$) are the parameters to estimate. For each observation, both the design matrix and the parameters $\bm \theta$ will change while the base density of $\bm n$ remain fixed.

We consider using elastic net regression for estimation of $\bm \theta$ as described in Appendix \ref{app:linear}, and the residual is calculated as
\beq \label{linres}
\bm{y}=\bm{x}-{\bf A}\widehat{\bmath \theta}.
\eeq

The calculation of ${\bm J}({\bm x},{\bm \theta})$ is also described in Appendix \ref{app:linear}. It may look plausible to combine multiple observations of $\bm x$ for estimating $\bm \theta$ to get an improved result. However, this is not possible because for each observation, we have a unique unknown value of $\bm \theta$.

In order to illustrate, we run a simulation with $D=15$ and $M=10$. We generate $90000$ samples of $\bm x$ with the following settings applied to each sample:
\begin{itemize}
  \item We generate $\bm A$ with entries drawn from $\mathcal{N}(0,1)$ and $\bm \theta$ with entries drawn from $\mathcal{U}(0,1)$.
  \item We generate the noise $\bm n$ with samples drawn from the multivariate Gaussian $\mathcal{N}({\bm \mu},\bm I)$ where $\bm \mu$ is a constant vector (initialized with entries drawn from $\mathcal{N}(0,0.1)$) for all simulations (note that $\| \bm \mu \| \approx 0$). We add the noise to the signal $\bm A \bm \theta$ by scaling the signal such that the signal to noise ratio is 1.
  \item We solve the elastic net regression problem to find the solution $\widehat{\bm \theta}$ as in (\ref{enet}). Using this solution, we find the residual $\bm y$ as in (\ref{linres}).
  \item The input to the DNN is $\bm y$ as well as metadata $\bm A$ and $\widehat{\bm \theta}$. The output is $\widetilde{\bm n}$, sampled from $p(\bm{n}; {\bmath \psi})$.
\end{itemize}

\begin{figure}
\begin{minipage}{0.98\linewidth}
\begin{center}
\centering
  \centerline{\includegraphics[width=1.0\textwidth]{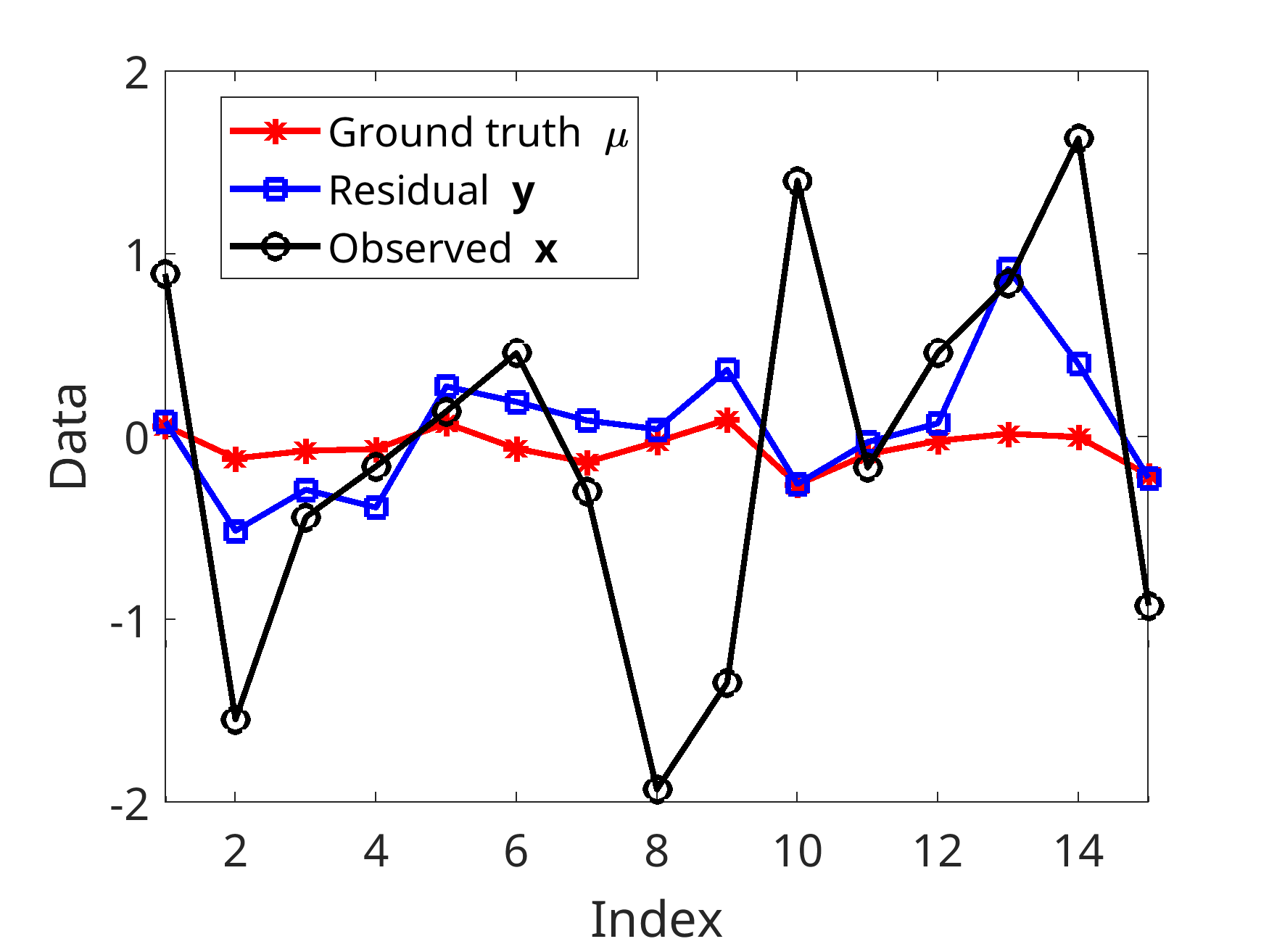}}
\end{center}
  \caption{An example simulation where the observed data $\bm x$, the residual data $\bm y$ and the ground truth weak signal $\bmath \mu$ are shown.
\label{data}}
\end{minipage}
\end{figure}

An example simulation is shown in Fig. \ref{data}. Note that the observed data $\bm x$ and the residual data $\bm y$ have higher power than the weak signal $\bm \mu$ that is hidden in noise. Even the residual $\bm y$ has higher power than $\bm \mu$, highlighting the need for averaging multiple observations of $\bm y$ to reach the weak signal.
We train our normalizing flow using the $90000$ samples and further details are given in Appendix \ref{app:training}.

\begin{figure}
\begin{minipage}{0.98\linewidth}
\begin{center}
\centering
  \centerline{\includegraphics[width=1.0\textwidth]{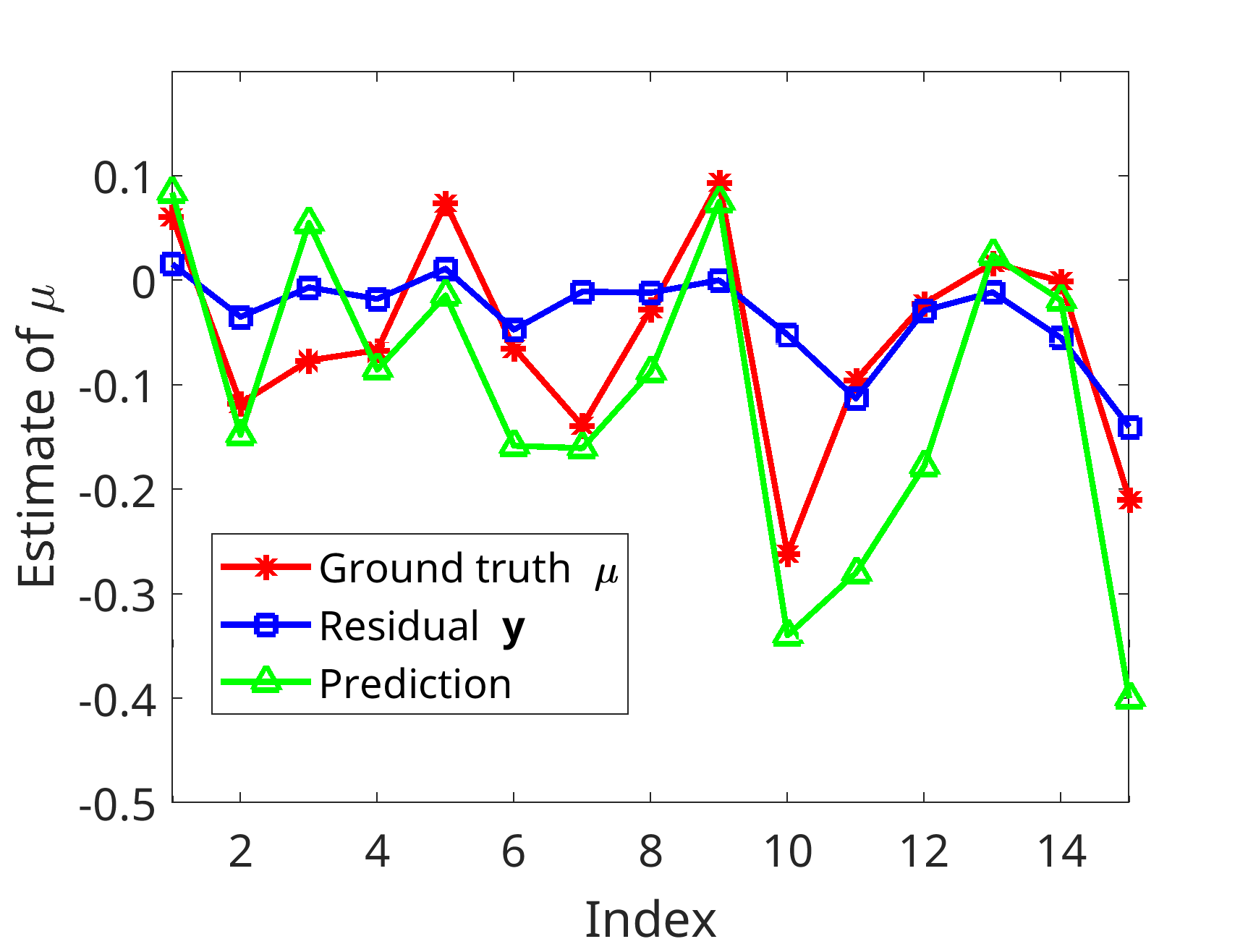}}
\end{center}
  \caption{Estimated weak signal using $3000$ samples compared with the ground truth (note the change in scale compared to Fig. \ref{data}). The averaged residual ${\bm y}$ gives a lower estimate of $\bm \mu$ (suppressing the signal and losing power). The average prediction by the trained normalized flow model $\widetilde{\bm n}$ gives a more accurate result.
\label{biases}}
\end{minipage}
\end{figure}

For evaluating the trained model, we simulate $3000$ additional samples using the same strategy as above. We average the residuals $\bm y$ of all $3000$ samples to find the estimates for $\bmath \mu$. For each simulation, we also do a prediction using our trained flow model which is also averaged. The results are shown in Fig. \ref{biases}. We see that while the averaged residual gives a suppressed result as the estimate of $\bm \mu$, the averaged prediction using the trained normalized flow gives a result more in agreement with the ground truth.

\subsection{Radio interferometric model\label{sec:radio}}
We present the data model used by a radio interferometer in this section. The observed data for baseline $pq$ at frequency $\nu$ and at any given time instance are given by ${\bm { V}}_{pq\nu}$ ($\in \mathbb{C}^{2\times 2}$) which is expressed as \citep{HBS}
\beq \label{vispq}
{\bm { V}}_{pq\nu}=\sum_{k \in [1,K]}  {\bm { J}}_{pk\nu} {\bm { C}}_{pqk\nu} {\bm { J}}_{qk\nu}^{H} + {\bm { N}}_{pq\nu}.
\eeq
The baseline $pq$ is formed by correlating the voltages received by stations $p$ and $q$. We consider the total number of stations (or receivers) to be $N$. The nuisance  signal in radio interferometry is generated by the sources in the sky (both Galactic and extra-Galactic). We model them as the summation of $K$ discrete signal as given by the first term on the right of (\ref{vispq}). Each signal in this summation, say the $k$-th, is a composition of the true signal from the sky, i.e., ${\bm { C}}_{pqk\nu}$ ($\in \mathbb{C}^{2\times 2}$) and terms representing systematic errors ${\bm { J}}_{pk\nu}, {\bm { J}}_{qk\nu}$ ($\in \mathbb{C}^{2\times 2}$) of receivers $p$ and $q$ along the path of the $k$-th source. The true source signal ${\bm { C}}_{pqk\nu}$  depends on the intensity of the source as well as its position in the sky ($l_k,m_k,n_k$ in radians) and the baseline coordinates in Fourier space ($u_{pq},v_{pq},w_{pq}$ in wavelengths). The weak signal of interest is in the noise ${\bm { N}}_{pq\nu}$ ($\in \mathbb{C}^{2\times 2}$) in (\ref{vispq}). The noise is assumed to be having a complex circular Gaussian density with zero mean and $\sigma^2 \bm I$ as covariance. However, due the presence of the unknown weak signal of interest, the true noise statistics are slightly different from the assumed statistics.

With calibration we estimate the systematic errors ${\bm { J}}_{pk\nu}$,${\bm { C}}_{pqk\nu}$ for all $p$,$q$,$k$ and $\nu$ (also with time dependence when data are taken over multiple sample instances). Thereafter, the residual ${\bf R}_{pq\nu}$ ($\in \mathbb{C}^{2\times 2}$) is extracted as 
\beq \label{residual}
{\bf R}_{pq\nu}={\bf V}_{pq\nu} - \sum_{k \in [1,K]} \widehat{\bm { J}}_{pk\nu} {\bm { C}}_{pqk\nu} \widehat{\bm { J}}_{qk\nu}^{H}. 
\eeq
where $\widehat{\bm { J}}_{pk\nu}$ and $\widehat{\bm { J}}_{qk\nu}$ represent the solutions obtained by calibration for the systematic errors. We can rewrite both (\ref{vispq}) and (\ref{residual}) in the form of (\ref{eq:obs}) and (\ref{eq:res}), respectively, by stacking the complex data as a vector of real components for all $p$,$q$ and $\nu$ and by considering the parameterization of ${\bm { J}}_{pk\nu}$ as a vector of real numbers in $\bmath \theta$.

In Appendix \ref{app:radio} we provide further details on calculating the influence function and (\ref{eq:jacobian}) for the radio interferometic data model. For this example, we set a simulation of a radio interferometric array with $N=14$ receivers, each receiver similar in instrumental response to a LOFAR high-band station \citep{LOFAR}. We generate $60000$ snapshot observations (with duration of one time sample with  $10$ s integration time and a bandwidth of 1 channel with $200$ kHz width) pointing at random locations in the sky  as training data. 
\begin{itemize}
  \item The beam pointing direction is randomly chosen and the epoch of observation is chosen so that the pointing direction is above the horizon.
  \item The nuisance signal is modeled as clusters of sources at $K=6$ distinct directions in the sky, at random separations from the pointing center. Their flux densities are randomly to be in $\mathcal{U}[0.1,200]$ Jy/PSF.
  \item The systematic errors ${\bm { J}}_{pk\nu}$ are generated to have entries drawn from a complex circular Gaussian distribution with zero mean and unit variance. In addition, an ${\bm I}$ is added to the diagonal of ${\bm { J}}_{pk\nu}$ for all $p$, $k$ and $\nu$.
  \item The frequency of the simulation is randomly chosen from $\mathcal{U}[110,150]$ MHz.
  \item The noise ${\bm N}_{pq}$ in (\ref{vispq}) is modeled by first drawing samples from a complex circular Gaussian distribution with zero mean and unit variance. To simulate the hidden weak signal, we generate a diffuse sky model that is kept fixed for all simulations and is centered at the north celestial pole. The diffuse sky model is simulated using a shapelet model (with random coefficients) as described in \citep{LSHAPELET}. Both these components are added together to create ${\bm N}_{pq}$. Note that the diffuse sky model has lower power compared to the noise component. Finally, the observed data ${\bm { V}}_{pq\nu}$ in (\ref{vispq}) are created by adding ${\bm N}_{pq}$ to the nuisance signal. The nuisance signal is scaled such that the nuisance signal to noise ratio is kept at $10$. Note that this is done so that the noise distribution $p(\bm{n})$ remain stable over all simulations.
  \item Direction dependent calibration is performed using the simulated data using specialized calibration software to get the residual (\ref{residual}), the same software also calculates the eigenvalues of the influence function (\ref{Reffderiv}) at no significant additional cost. Note that we assume perfect knowledge of the $K=6$ sky model (nuisance signal) but no information of the diffuse sky is used during calibration.
  \item For training the DNN, we use the residual ${\bf R}_{pq\nu}$ (\ref{residual}) as input (in vectorized form as ${\bm y}$). For $N=14$ we have $D=N(N-1)/2\times 8=728$ values as input. We use the following as metadata: The sky model of the nuisance signal of $K=6$ source clusters. The calibration solutions ${\widehat{\bm { J}}_{pk\nu}}$ for all $p$ and $k$. The observed data ${\bf V}_{pq\nu}$ for all $p$ and $q$. Finally, the coordinates of the baselines in Fourier space, i.e., all $u_{pq},v_{pq},w_{pq}$ triplets for all $p$ and $q$.
  \item The output of the DNN is a prediction of the noise signal $\widehat{\bm { N}}_{pq\nu}$ with the weak diffuse emission hopefully intact.
\end{itemize}

In Appendix \ref{app:training}, we provide further practical details about the DNN model and the training strategy. After training the DNN, we evaluate its performance by generating a separate test observation. We generate an observation with $2$ hr duration ($10$ s integration time), with $20$ frequencies equally spaced in the range $[110,150]$ MHz. The remaining details of the simulation are similar to the settings used in the generation of training data as described previously.

\begin{figure*}
\begin{minipage}{1.00\linewidth}
\begin{center}
\begin{minipage}{0.24\linewidth}
\centering
  \centerline{\includegraphics[width=1.0\textwidth]{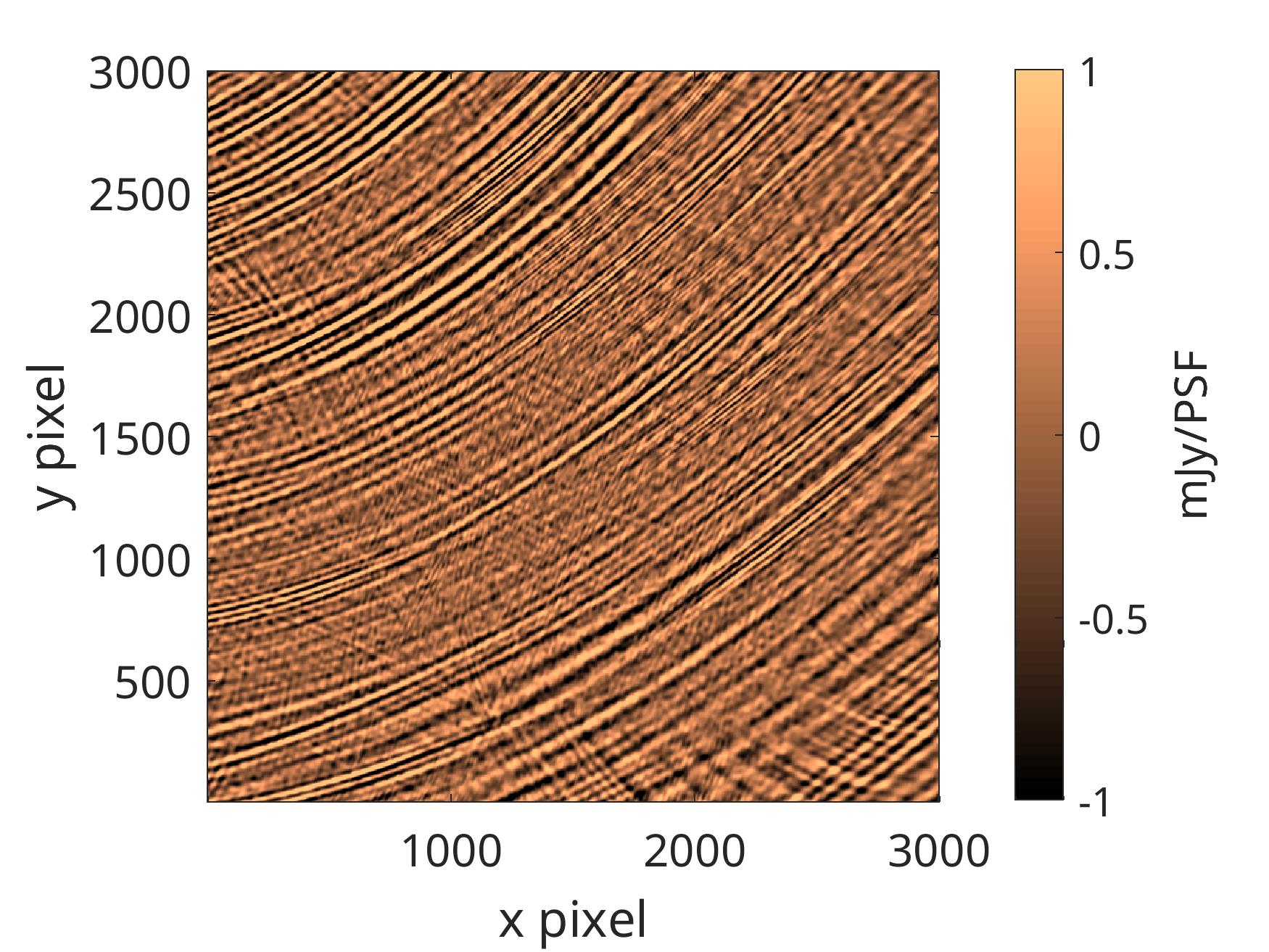}}
\vspace{0.1cm}
\end{minipage}
\begin{minipage}{0.24\linewidth}
\centering
 \centerline{\includegraphics[width=1.0\textwidth]{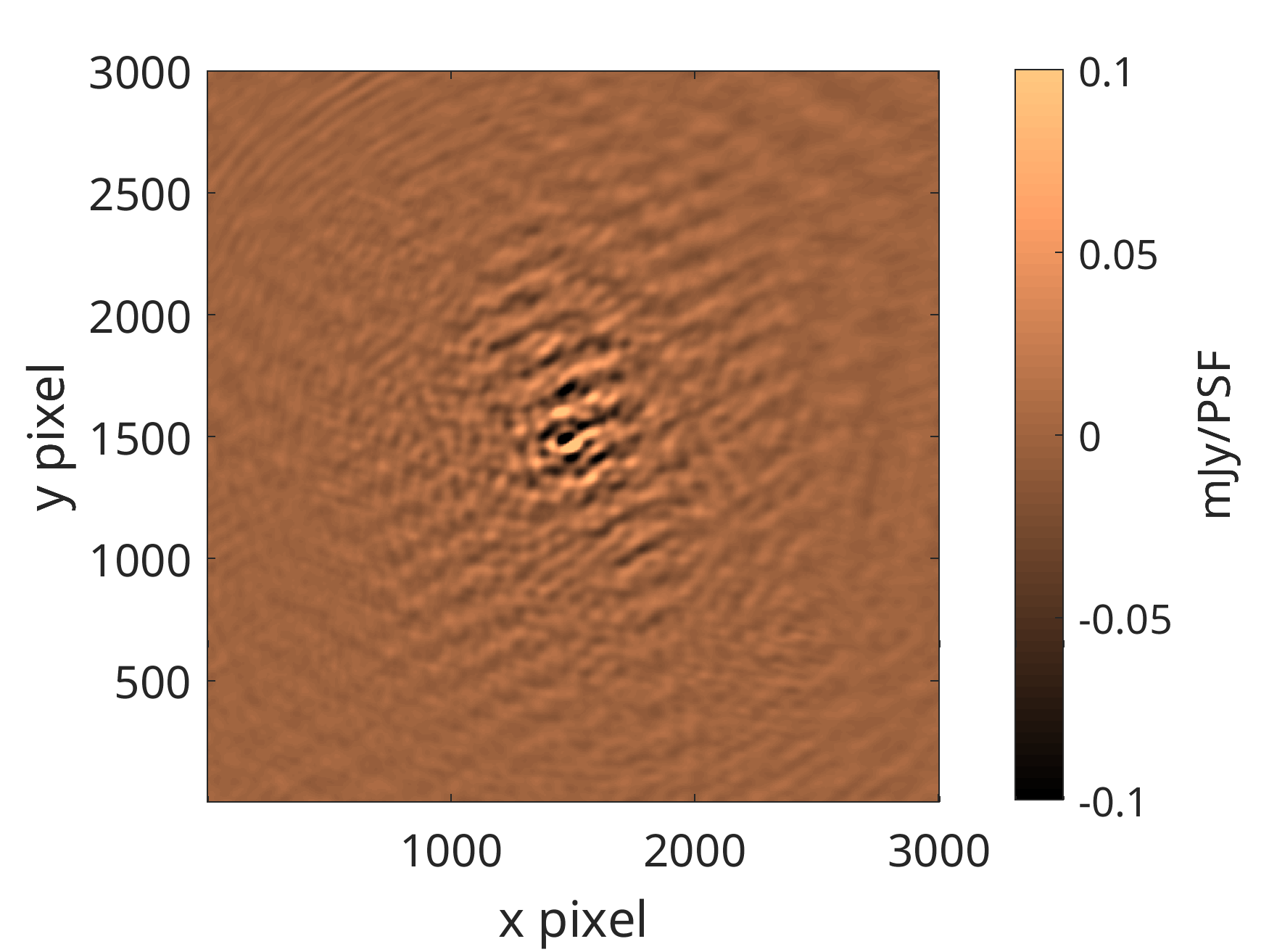}}
\vspace{0.1cm}
\end{minipage}
\begin{minipage}{0.24\linewidth}
\centering
 \centerline{\includegraphics[width=1.0\textwidth]{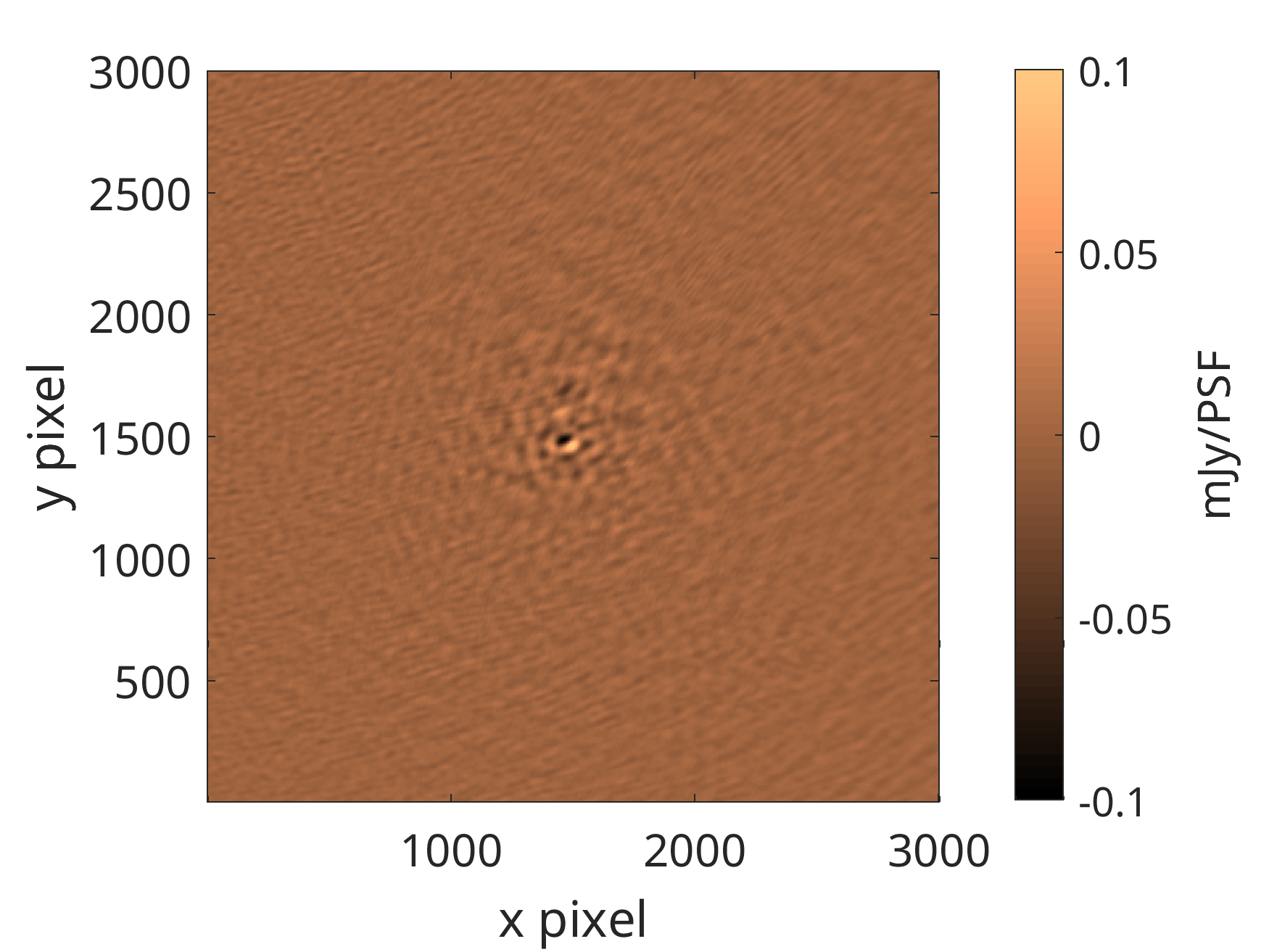}}
\vspace{0.1cm}
\end{minipage}
\begin{minipage}{0.24\linewidth}
\centering
 \centerline{\includegraphics[width=1.0\textwidth]{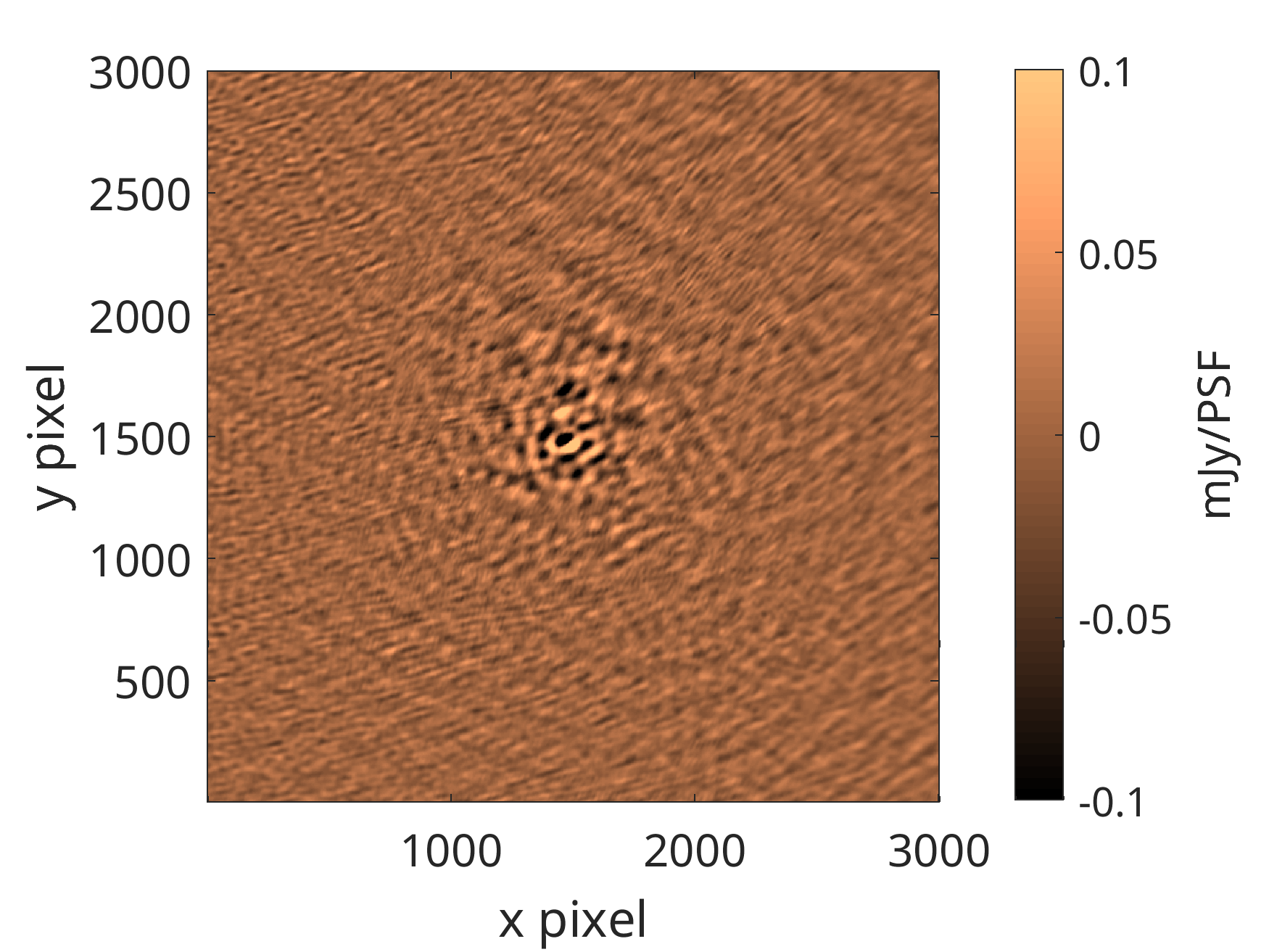}}
\vspace{0.1cm}
\end{minipage}\\
\begin{minipage}{0.24\linewidth}
\centering
  \centerline{\includegraphics[width=1.0\textwidth]{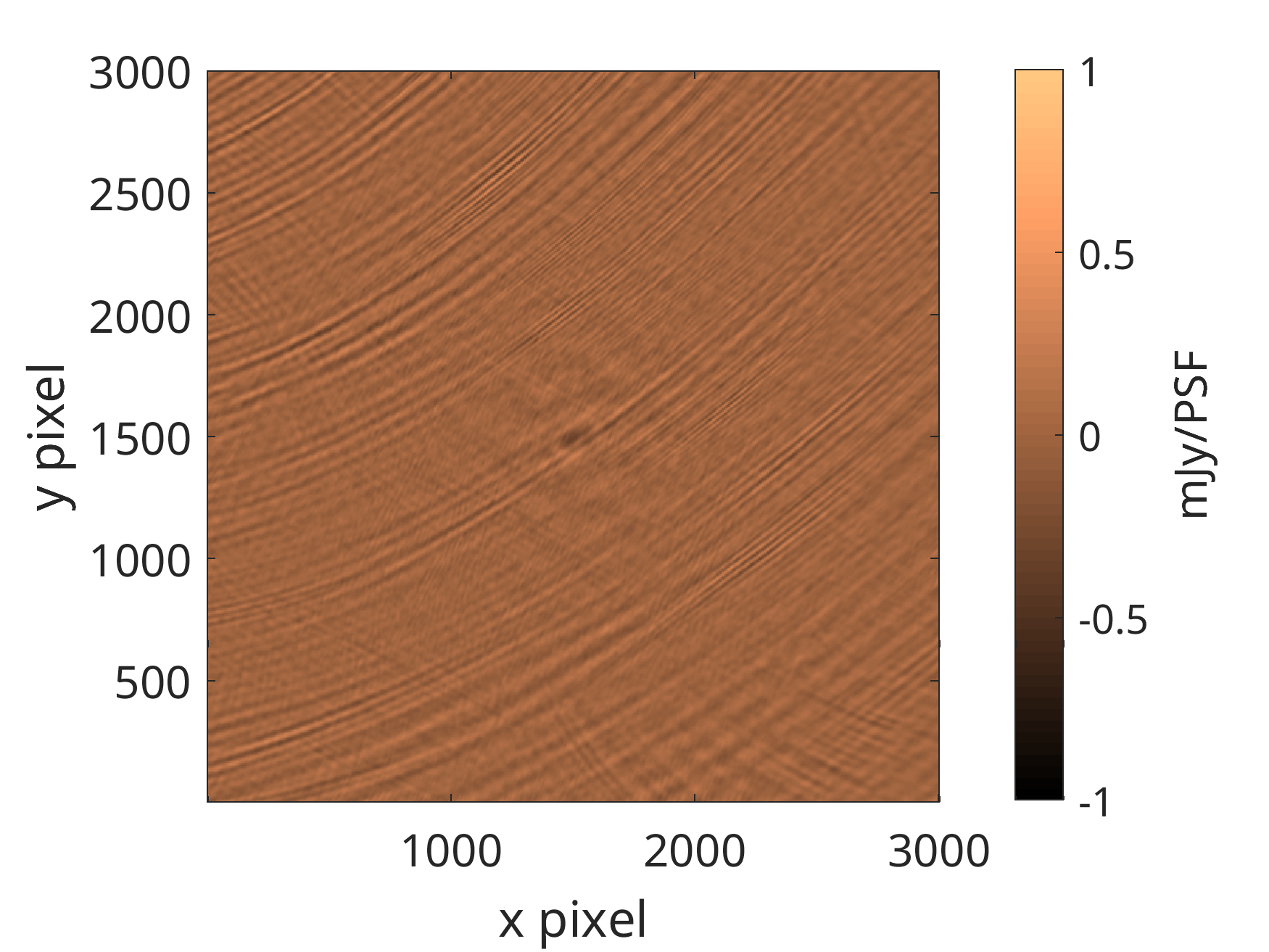}}
\vspace{0.1cm}
\end{minipage}
\begin{minipage}{0.24\linewidth}
\centering
 \centerline{\includegraphics[width=1.0\textwidth]{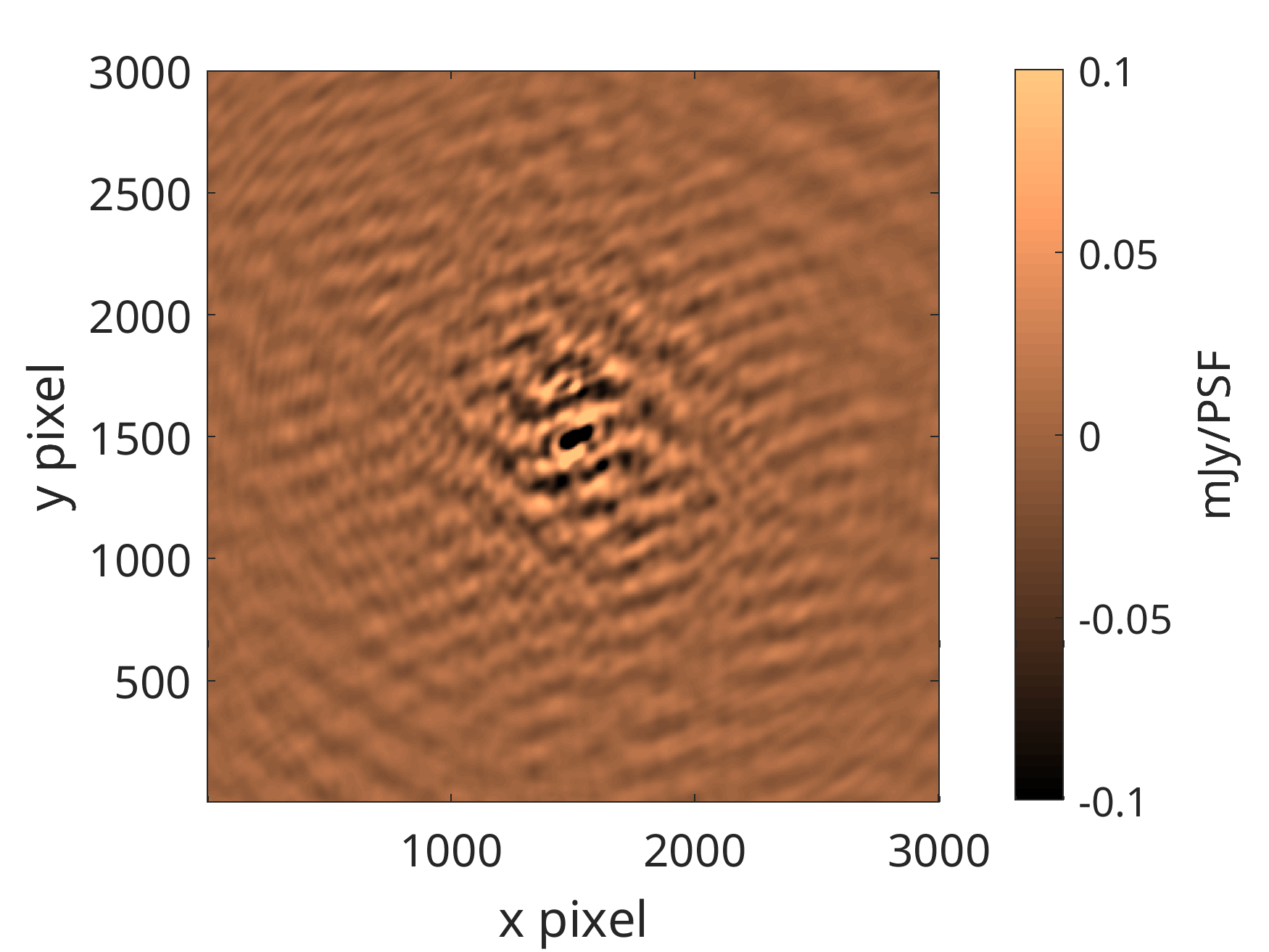}}
\vspace{0.1cm}
\end{minipage}
\begin{minipage}{0.24\linewidth}
\centering
 \centerline{\includegraphics[width=1.0\textwidth]{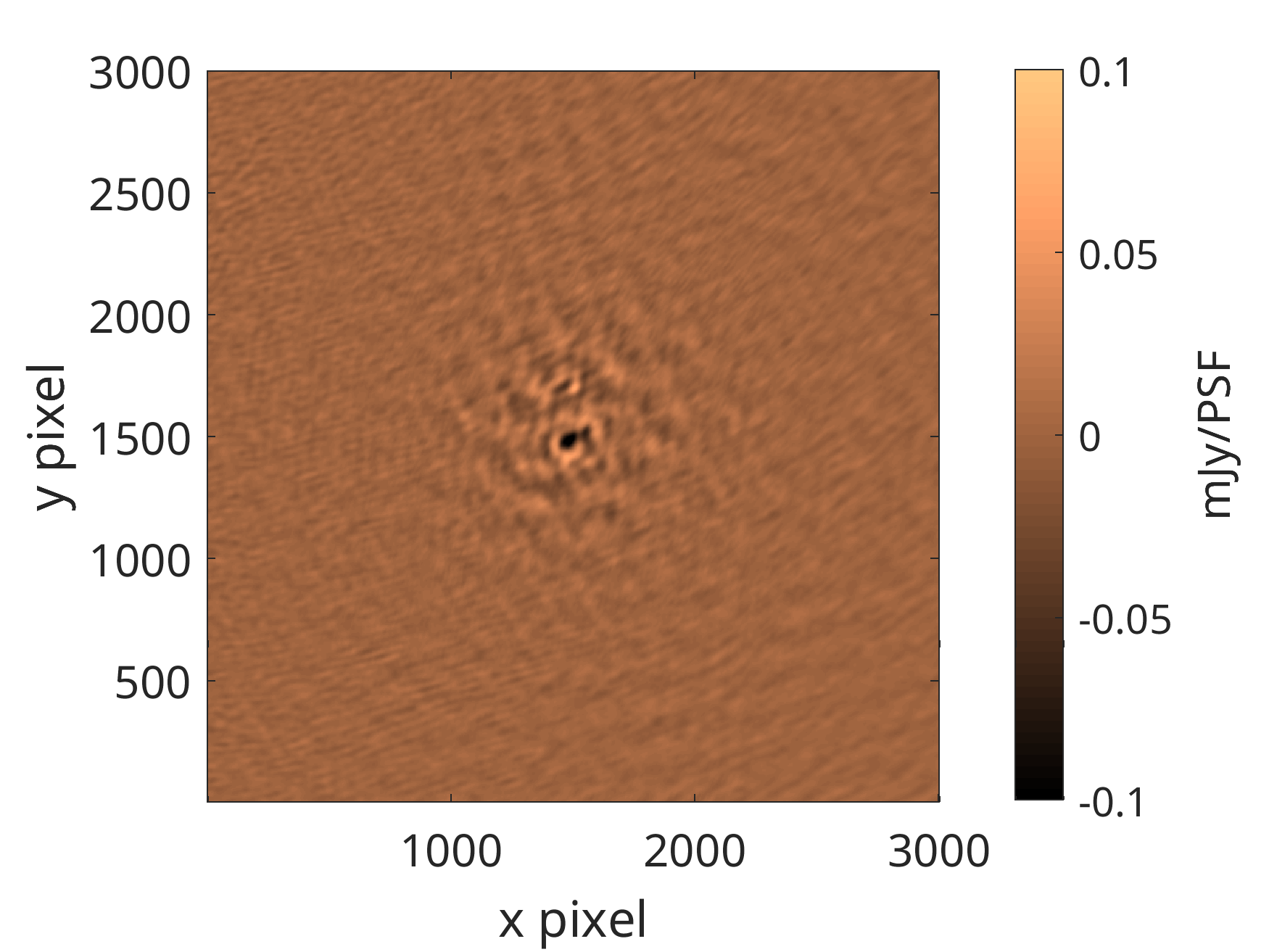}}
\vspace{0.1cm}
\end{minipage}
\begin{minipage}{0.24\linewidth}
\centering
 \centerline{\includegraphics[width=1.0\textwidth]{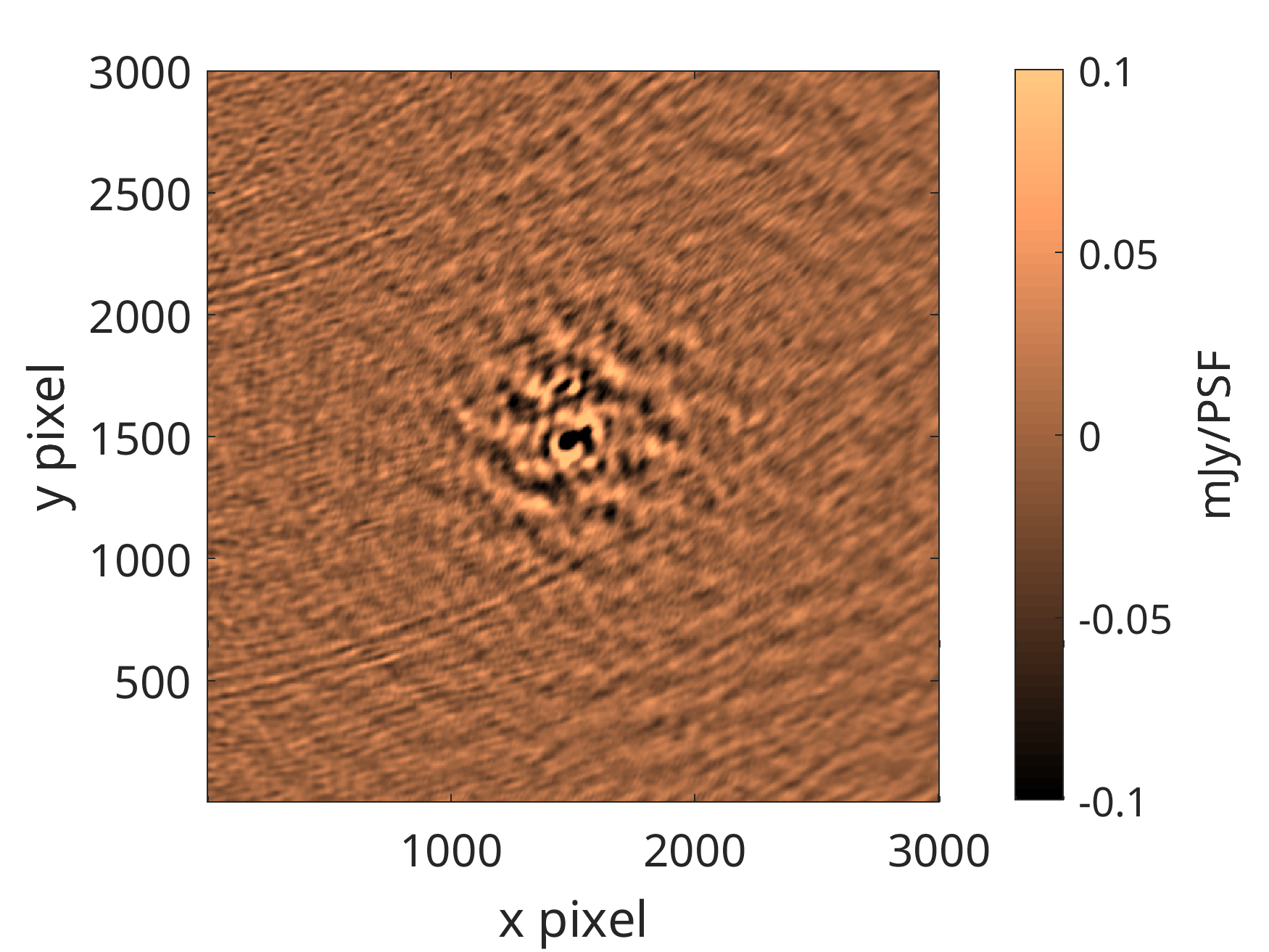}}
\vspace{0.1cm}
\end{minipage}\\
\begin{minipage}{0.24\linewidth}
\centering
  \centerline{\includegraphics[width=1.0\textwidth]{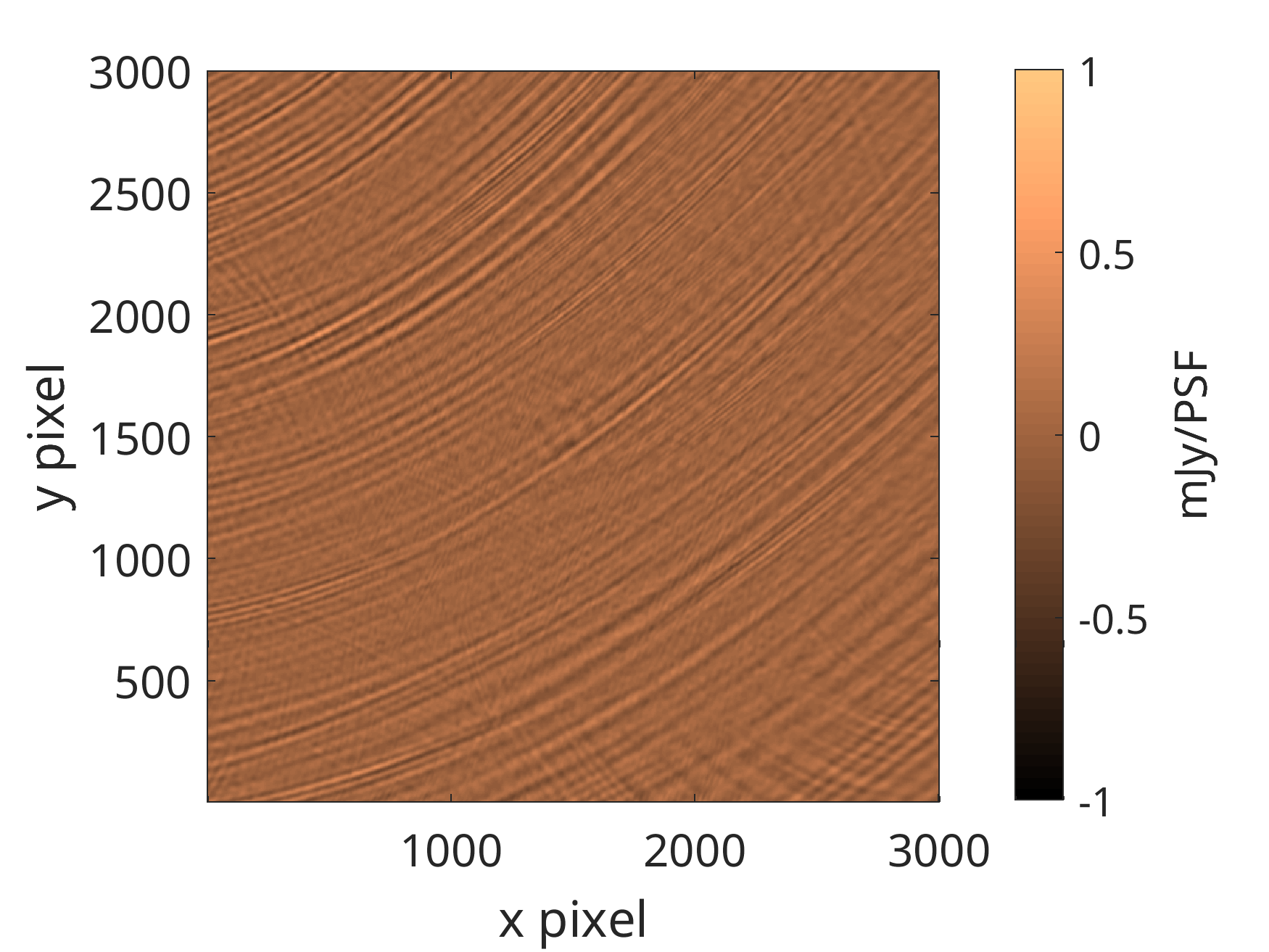}}
\vspace{0.1cm}
\end{minipage}
\begin{minipage}{0.24\linewidth}
\centering
 \centerline{\includegraphics[width=1.0\textwidth]{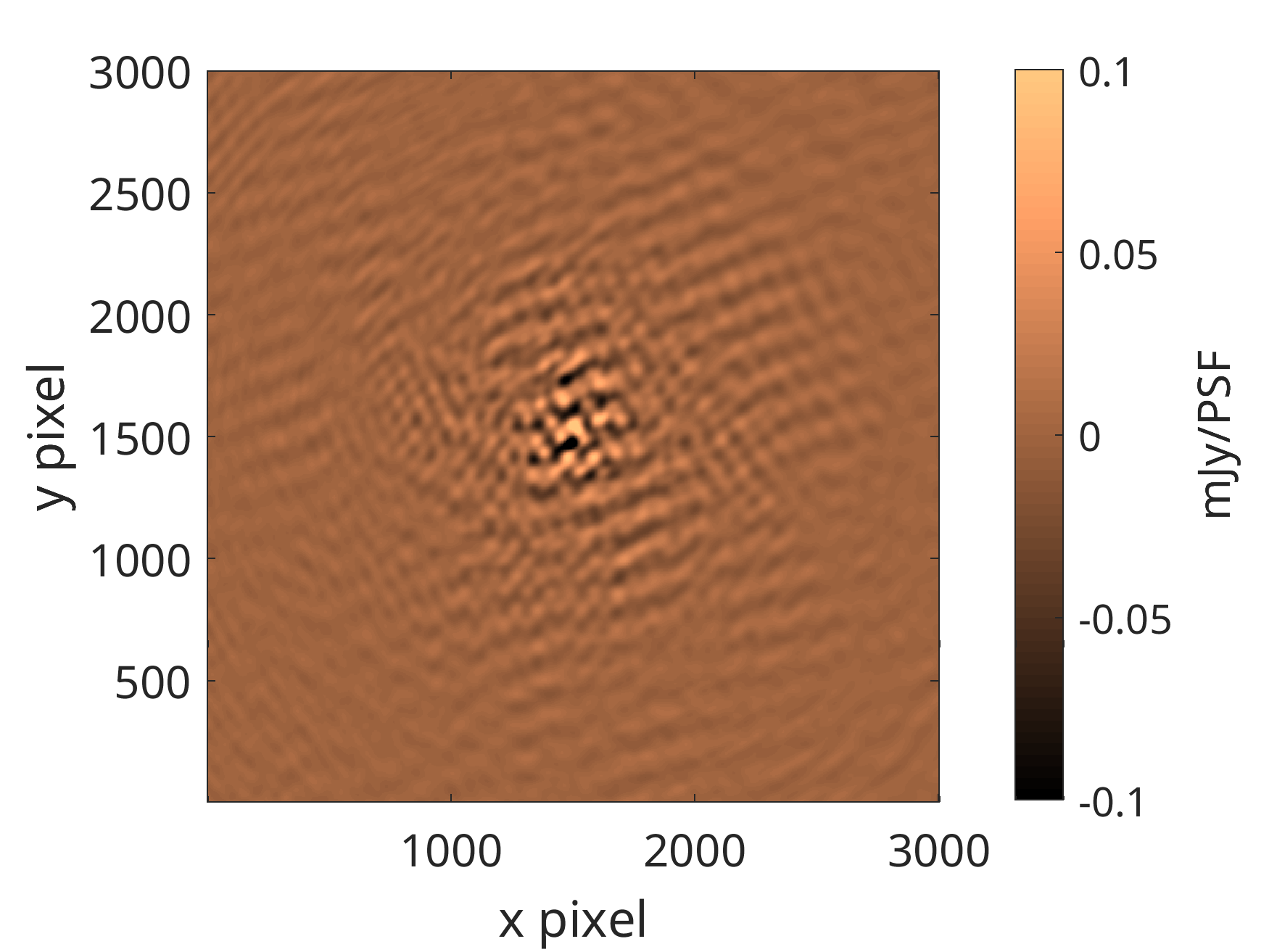}}
\vspace{0.1cm}
\end{minipage}
\begin{minipage}{0.24\linewidth}
\centering
 \centerline{\includegraphics[width=1.0\textwidth]{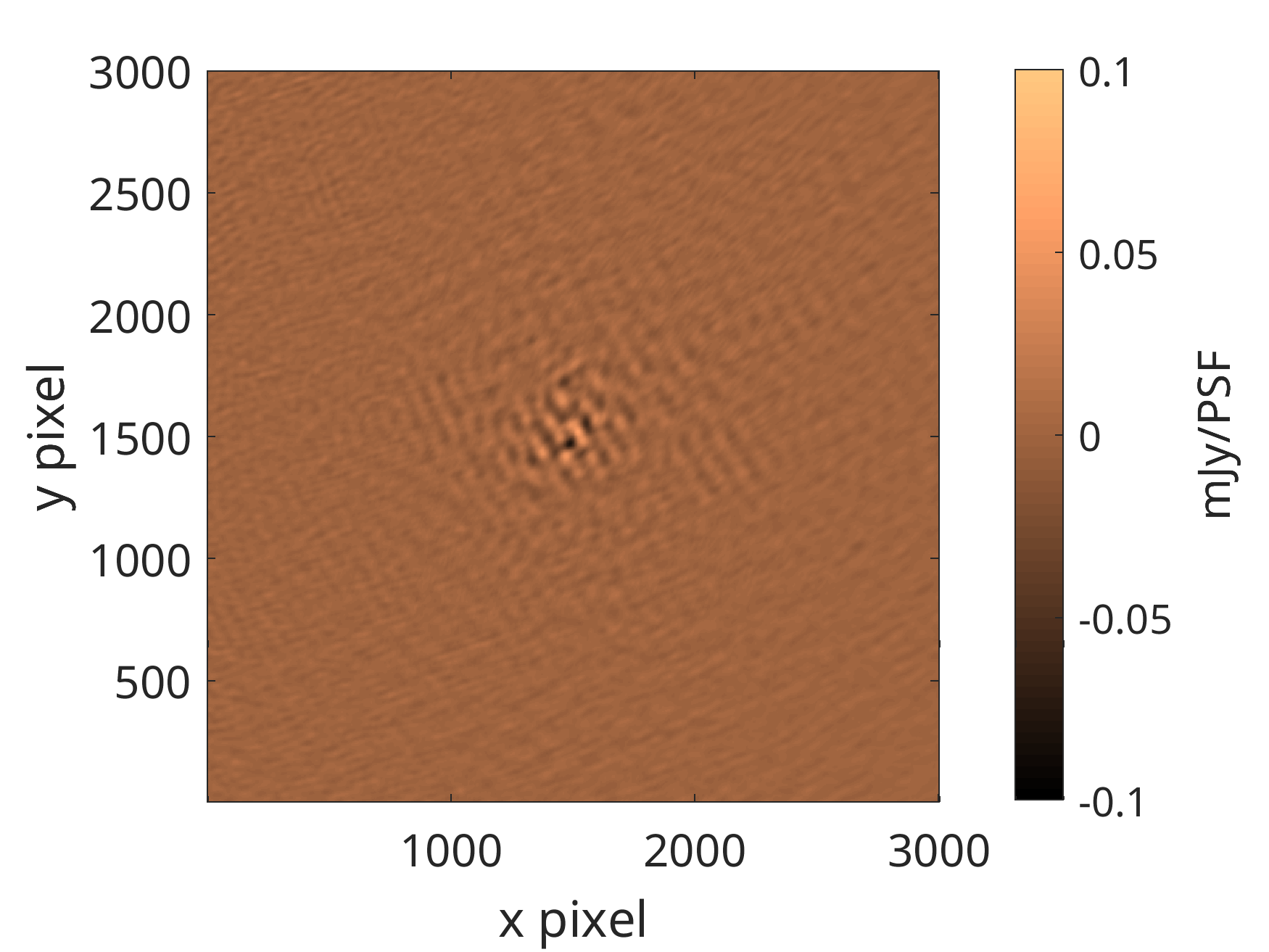}}
\vspace{0.1cm}
\end{minipage}
\begin{minipage}{0.24\linewidth}
\centering
 \centerline{\includegraphics[width=1.0\textwidth]{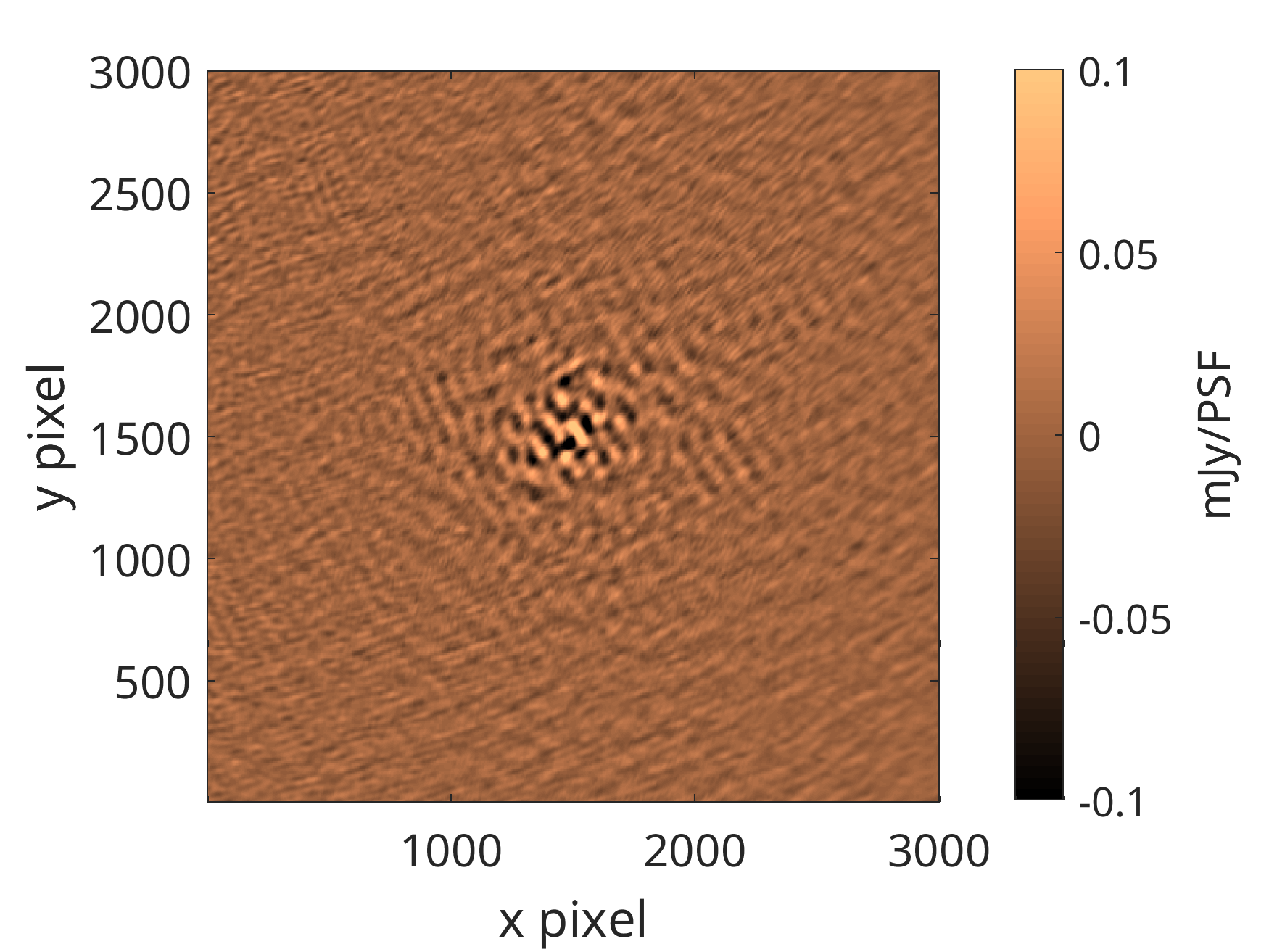}}
\vspace{0.1cm}
\end{minipage}\\
\begin{minipage}{0.24\linewidth}
\centering
  \centerline{\includegraphics[width=1.0\textwidth]{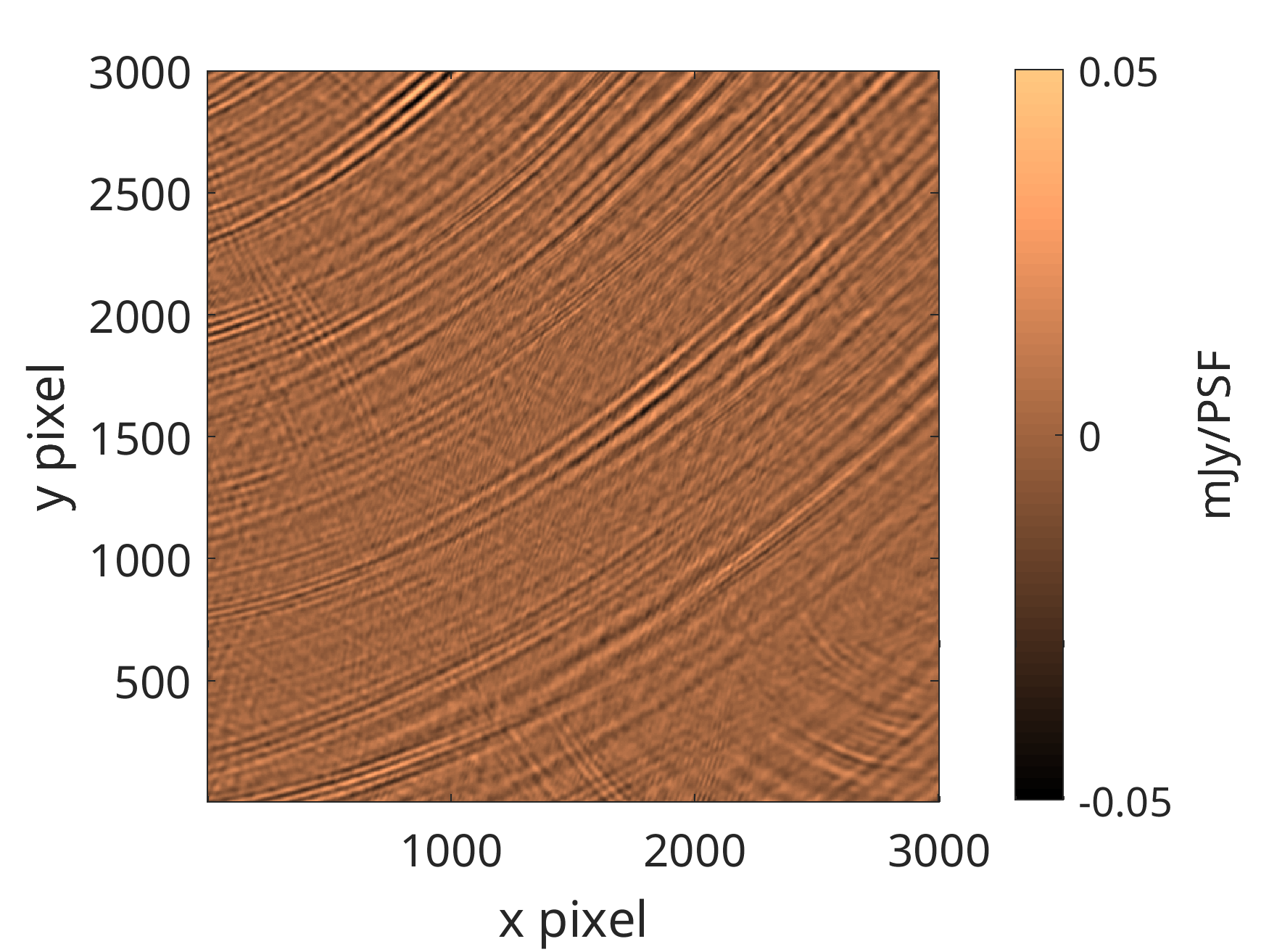}}
\vspace{0.5cm} \centerline{Data}\smallskip
\end{minipage}
\begin{minipage}{0.24\linewidth}
\centering
 \centerline{\includegraphics[width=1.0\textwidth]{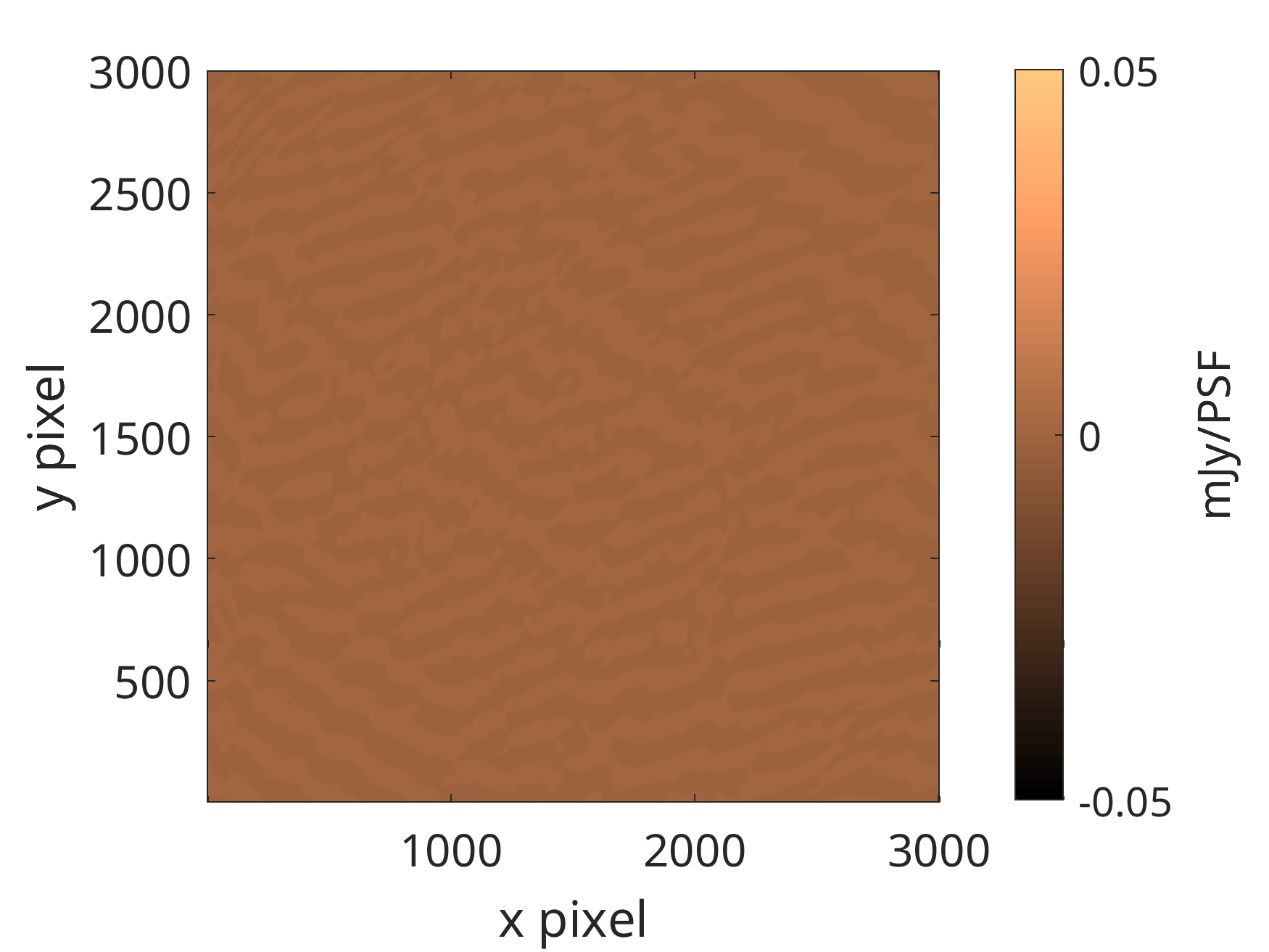}}
\vspace{0.5cm} \centerline{Ground truth}\smallskip
\end{minipage}
\begin{minipage}{0.24\linewidth}
\centering
 \centerline{\includegraphics[width=1.0\textwidth]{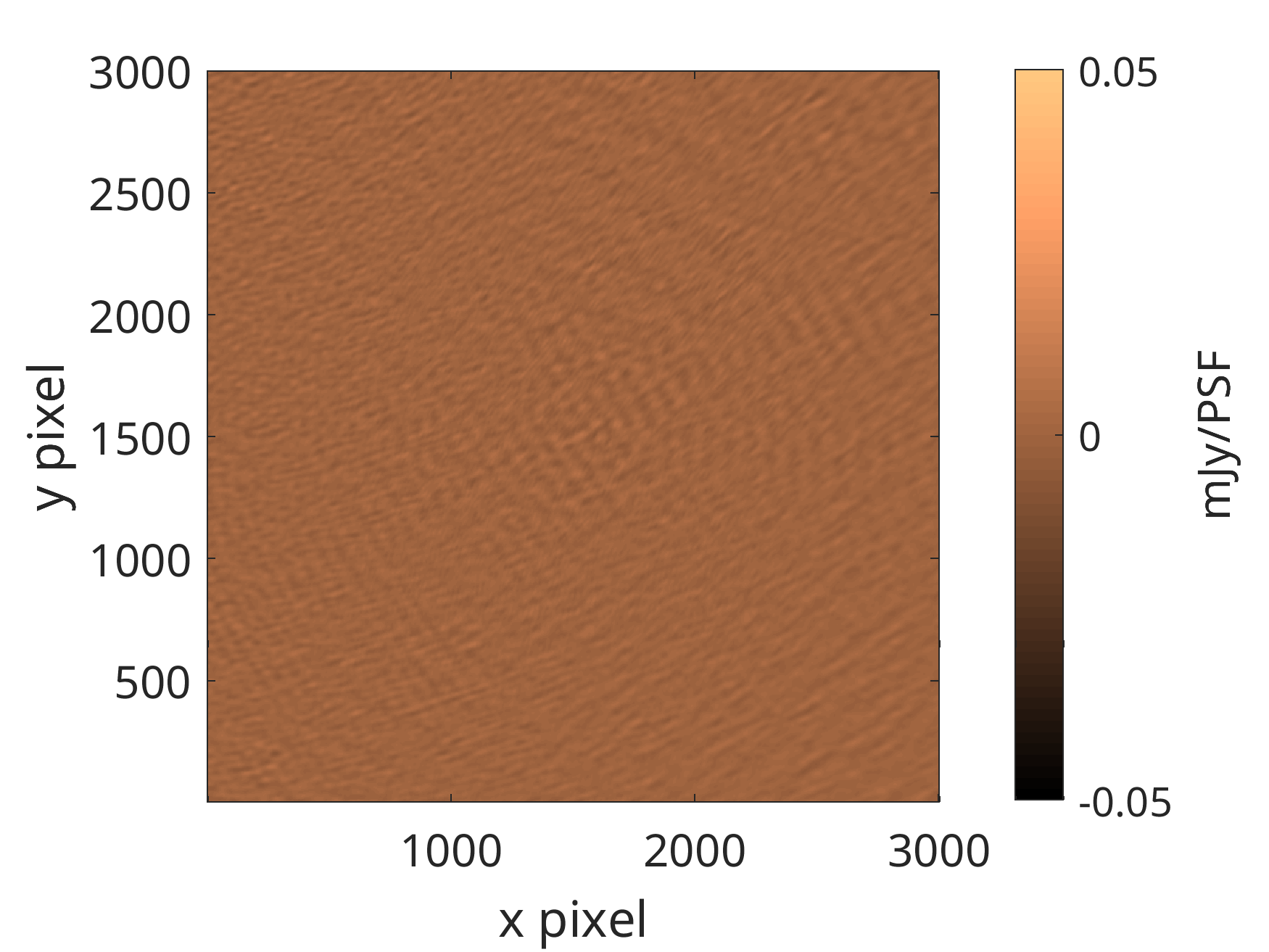}}
\vspace{0.5cm} \centerline{Residual}\smallskip
\end{minipage}
\begin{minipage}{0.24\linewidth}
\centering
 \centerline{\includegraphics[width=1.0\textwidth]{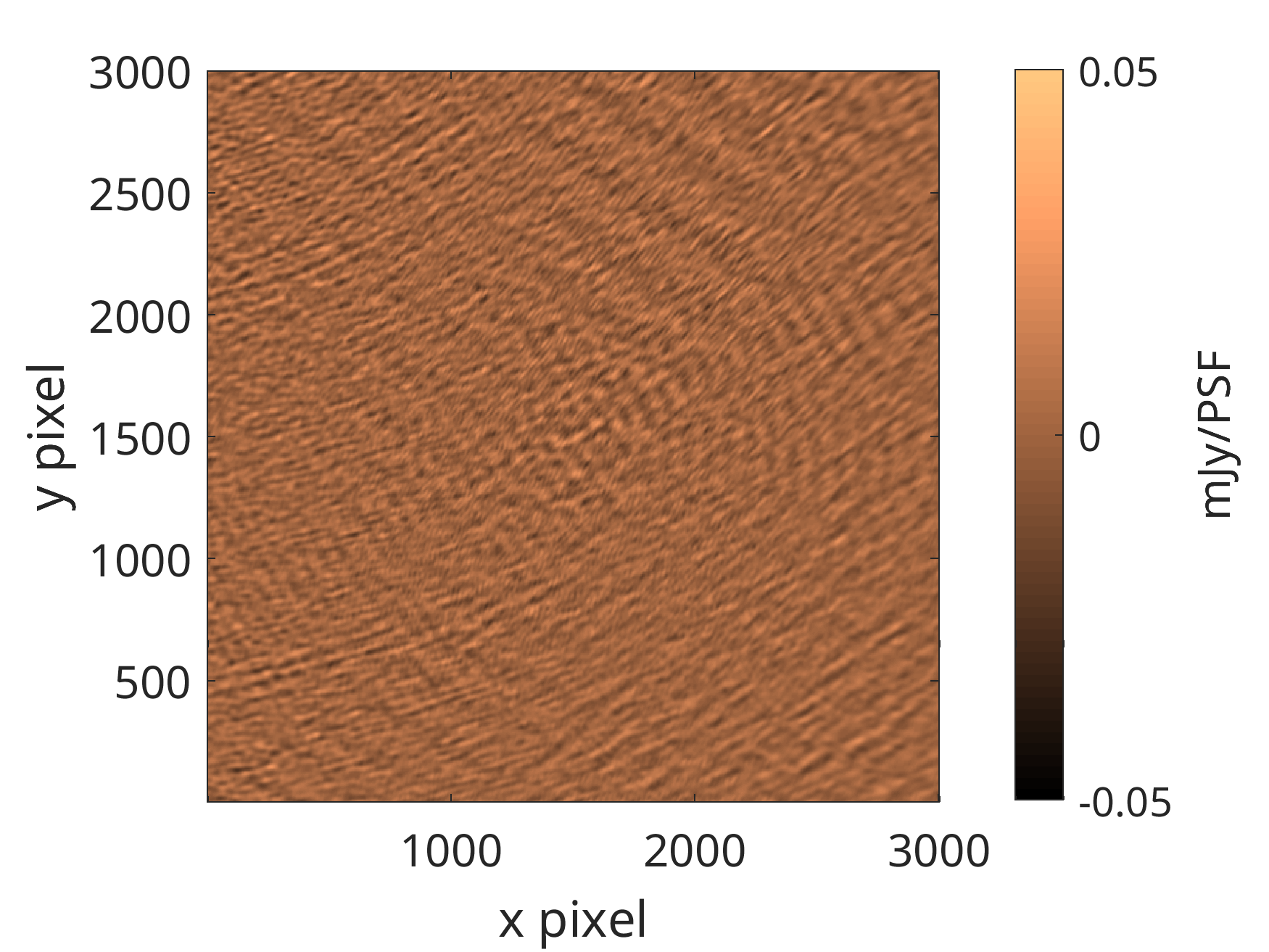}}
\vspace{0.5cm} \centerline{Prediction}\smallskip
\end{minipage}\\
\end{center}
\caption{
  Average images made using the test observation (all frequencies), covering about $8 \times 8$ square degrees in the sky, centered at the north celestial pole. Note that the beam pointing direction of this observation is elsewhere in the sky while the weak (diffuse) signal is simulated to be at the pole. The four columns from left to right show the images made using the observed data (prior to calibration), the ground truth (without noise), the residual (after calibration) and the prediction from the DNN, respectively. Note the difference in the color scale in the left column, where the dominant component is the nuisance signal and its artifacts. The four rows from top to bottom correspond to the Stokes I, Q, U and V polarizations. Note that the images made using the residual has significant suppression of the desired signal while the DNN prediction closely agrees with the ground truth. The ground truth images are without any noise while the DNN prediction contains the expected noise as well. This is evidenced by the Stokes V images in the last row, where the image made by uncalibrated data (leftmost column) and the DNN prediction (rightmost column) have more or less the same power level.
\label{maps_all}}
\end{minipage}
\end{figure*}

In Fig. \ref{maps_all}, we show images made using the test observation. Data at all frequencies are imaged separately and average to get the final image. The calibration is performed per each time and frequency sample and the DNN model prediction is also done at this resolution. One noteworthy aspect in the DNN model as opposed to the DNN model trained in the linear example is the use of Monte Carlo dropout \citep{Gal2015}. Therefore, multiple predictions are performed at each time and frequency sample and their average is taken as the final result.

From the images in Fig. \ref{maps_all} we clearly see that the DNN prediction does a more accurate prediction of the hidden signal compared to just using the residual, closely matching the ground truth. Note that the ground truth images do not show any noise (because only the diffuse sky model is simulated).

\section{Conclusions\label{sec:conc}}
We have presented the use of normalizing flows for recovering weak signals that are affected by calibration. Using simulated data for training and testing the normalizing flow model, we have shown the improved reconstruction of the weak signal. Note that with real observations, we do not need any simulated data for training the DNN as we can directly feed the observed data and metadata into the training process. Specifically for radio interferometry, by using this method, we can combine various observations done aimed at diverse science goals and even observations done by various radio telescopes to probe deeper into the Universe.

Data Availability Statement: All data used in this paper are simulated and the software used for simulations and evaluations are provided at \url{https://github.com/SarodYatawatta/InfluenceFlow}.


\appendix

\section{Proof of (\ref{eq:jacobian}) \label{app:influence}}
We give the proof based on prior work \cite{Gould2016,Samuel,SAM2018}.
Let $x_m$ be the $m$-th element of $\bm{x}$, we take the derivative of (\ref{eq:res})
\beq
\frac{\partial \bm{ y}}{\partial x_m} = \frac{\partial \bm{ x}}{\partial x_m} - \frac{\partial}{\partial x_m}s(\bm{\theta})\arrowvert_{\bm{\theta}=\widehat{\bm{\theta}}}.
\eeq

Using the chain rule (at ${\bm{\theta}}=\widehat{\bm{\theta}}$)
\beq \label{s_der}
 \frac{\partial}{\partial x_m} {\bf s}({\bmath \theta}) \arrowvert_{{\bmath \theta}=\widehat{\bmath \theta}} =\frac{\partial {\bf s}({\bmath \theta})}{\partial {\bmath \theta}^T} \arrowvert_{{\bmath \theta}=\widehat{\bmath \theta}} \times \frac{\partial \widehat{\bmath \theta}}{\partial x_m}
\eeq
where $\frac{\partial {\bf s}({\bmath \theta})}{\partial {\bmath \theta}^T}  \in \mathbb{R}^{D\times M}$ and $\frac{\partial {\bmath \theta}}{\partial x_m} \in \mathbb{R}^{M\times 1}$.

At the solution of (\ref{eq:mle}), we have a local minimum, i.e.,  $\frac{\partial f({\bf x},{\bmath \theta})}{\partial {\bmath \theta}} \arrowvert_{{\bmath \theta}=\widehat{\bmath \theta}}  ={\bf 0}$. Let $f^{\prime}({\bf x},{\bmath \theta})=\frac{\partial f({\bf x},{\bmath \theta})}{\partial {\bmath \theta}}$ for simplified notation. Then $f^{\prime}({\bf x},\widehat{\bmath \theta})={\bf 0}$ and taking derivative of both sides with respect to $x_m$, we get $\frac{\partial f^\prime}{\partial x_m} \frac{\partial x_m}{\partial x_m}+ \frac{\partial f^\prime}{\partial \widehat{\bmath \theta}} \frac{\partial \widehat{\bmath \theta}}{\partial x_m} ={\bf 0}$. Simplifying this leads to (\ref{theta_der}),
\beq \label{theta_der}
\frac{\partial \widehat{\bmath \theta}}{\partial x_m} = - \left( f_{\theta \theta} ({\bf x},{\bmath \theta}) \right)^{-1} f_{x_m \theta} ({\bf x},{\bmath \theta}) \arrowvert_{{\bmath \theta}=\widehat{\bmath \theta}}
\eeq
where
\beqn
f_{\theta \theta} ({\bf x},{\bmath \theta}) \buildrel \triangle \over= \frac{\partial^2 f({\bf x},{\bmath \theta})} {\partial {\bmath \theta} \partial{\bmath \theta}^T}   \in \mathbb{R}^{M\times M},\\
f_{x_m \theta} ({\bf x},{\bmath \theta}) \buildrel \triangle \over= \frac{\partial^2 f({\bf x},{\bmath \theta})}{\partial x_m \partial {\bmath \theta}}  \in \mathbb{R}^{M\times 1}.
\eeqn

Taking the derivative of (\ref{eq:res}) with respect to $\bm{ x}$, we get ${\bm I}$ for the first term and for the second term, we use (\ref{s_der}) and (\ref{theta_der}) above to get (\ref{eq:jacobian}).

\section{Linear model \label{app:linear}}
With elastic net regression, the estimate $\widehat{{\bmath \theta}}$ is obtained as
\beq \label{enet}
\widehat{{\bmath \theta}}=\underset{\bmath \theta}{\argmin}\left(\|{\bf x}-{\bf A}{\bmath \theta}\|^2 + \rho_2 \|{\bmath \theta}\|^2+ \rho_1 \|{\bmath \theta}\|_1 \right)
\eeq
where $\rho_1$ and $\rho_2$ are regularization factors (both set at $0.001$).
The influence function is obtained as $\bm{I}+\mathcal{A}$ where we use (\ref{eq:jacobian}) to get
\beq \label{calAA}
\mathcal{A} = {\bf A}\frac{1}{2}\left({\bf A}^T{\bf A}+(\rho_2+\rho_1 \delta(\|{\bmath \theta}\|)) {\bf I}\right)^{-1}\left(-2{\bf A}^T\right)
\eeq
where $\delta(\cdot)$ is the Dirac delta function.

\section{Radio interferometric model \label{app:radio}}
In order to simplify the notation, we introduce the canonical selection matrix ${\bm A}_p$ ($\in \mathbb{R}^{2\times 2N}$), where only the $p$-th block is ${\bf I} \in \mathbb{R}^{2\times 2}$, the remaining entries are $0$. With this notation, we have for example, ${\bm { J}}_{pk\nu} = {\bm A}_p {\bm J}_{k\nu}$ where ${\bm J}_{k\nu}$ ($\in \mathbb{C}^{2N\times 2}$) is a block matrix collecting systematic errors for all $N$ stations along the $k$-th direction at frequency $\nu$ into one. We minimize the cost function
\beqn \label{cost}
\lefteqn{g_{\nu}({\bm J}_{k\nu};k\in[1,K])=}\\\nonumber
&&\sum_{p,q}\| {\bm V}_{pq\nu} - \sum_{k \in [1,K]} {\bm A}_p{\bm J}_{k\nu} {\bm C}_{pqk\nu} ({\bm A}_q{\bm J}_{k\nu})^H \|^2
\eeqn
to find the solutions ${\bm J}_{k\nu}$ for all $k$ and $\nu$, using data at all baselines (all possible $p$ and $q$, which is $N(N-1)/2$ for one frequency).

Consider $x_{p^\prime q^\prime r}$ to be an element in the observed data vector $\bm x$ corresponding to the baseline formed by stations $p^\prime$ and $q^\prime$, and $r\in [0,7]$ is the index to denote which component of the $2\times 2$ matrix it represents. We take the derivative of (\ref{residual}) by $x_{p^\prime q^\prime r}$ to get
\beqn \label{Reffderiv}
 \mathrm{vec}\left( \frac{\partial{{\bm R}}_{pq\nu}} {\partial x_{p^\prime q^\prime r}} \right)&& =\mathrm{vec}\left( \frac{\partial{\bm V}_{pq\nu}} {\partial x_{p^\prime q^\prime r}} \right)\\\nonumber
-\sum_{k \in [1,K]} && \left({\bm C}_{pqk\nu} \widehat{\bm J}_{k\nu}^H {\bm A}_q^T\right)^T \otimes {\bm A}_p \mathrm{vec}\left( \frac{\partial{\bm J}_{k\nu}} {\partial x_{p^\prime q^\prime r} } \right)
\eeqn
where analogous to (\ref{theta_der}), we have \cite{SS5}
\beqn \label{Jderiv}
\lefteqn{\mathrm{vec}\left(\frac{\partial {\bm J}_{k\nu}}{\partial x_{p^\prime q^\prime r}}\right)}\\\nonumber
&&=\left( \mathcal{D}_{\bm J}{\rm grad}_k(g_{\nu}({\bm J}_{k^\prime\nu};k^\prime\in[1,K]))\right)^{-1} \\\nonumber
&&\times \left({\bm A}_{q^\prime}{\bm J}_{k\nu} {\bm C}_{p^\prime q^\prime k\nu}^H\right)^T \otimes {\bm A}_{p^\prime}^T \mathrm{vec}\left(\frac{\partial {\bm V}_{p^\prime q^\prime \nu}}{\partial x_{p^\prime q^\prime r}}\right)
\eeqn
and
\beqn \label{DJ}
\lefteqn{\mathcal{D}_{\bm J}{\rm grad}_k(g_{\nu}({\bm J}_{k^\prime\nu};k^\prime\in[1,K]))=}\\\nonumber
&&\sum_{p,q} \left( -({\bm C}_{pqk\nu}^{H})^{T}\otimes {\bm A}_p^T {\bm R}_{pq\nu}{\bm A}_q
-{\bm C}_{pqk\nu}^{T}\otimes {\bm A}_q^T {\bm R}_{pq\nu}^H{\bm A}_p  \right. \\\nonumber
&+&\left.({\bm C}_{pqk\nu}{\bm J}_{k\nu}^H{\bm A}_q^T{\bm A}_q{\bm J}_{k\nu}{\bm C}_{pqk\nu}^H)^T\otimes {\bm A}_p^T{\bm A}_p \right. \\\nonumber
&+&\left.({\bm C}_{pqk\nu}^H{\bm J}_{k\nu}^H{\bm A}_p^T{\bm A}_p{\bm J}_{k\nu}{\bm C}_{pqk\nu})^T\otimes {\bm A}_q^T{\bm A}_q \right).
\eeqn

Finally, we rearrange (\ref{Reffderiv}) in vectorized form, with real and imaginary parts of each $\mathbb{C}^{2\times 2}$ matrix represented by 8 real values in the vector $\bm x$ for any given $p$,$q$ and $\nu$. Adding spectral regularization as in \citep{DCAL}, modifies the above formulae as in \citep{ST2019} but we leave them out here for simplicity.

\section{Deep learning model and training details\label{app:training}}
\begin{figure}[ht]
  \begin{minipage}{0.99\linewidth}
    \begin{center}
  \epsfig{figure=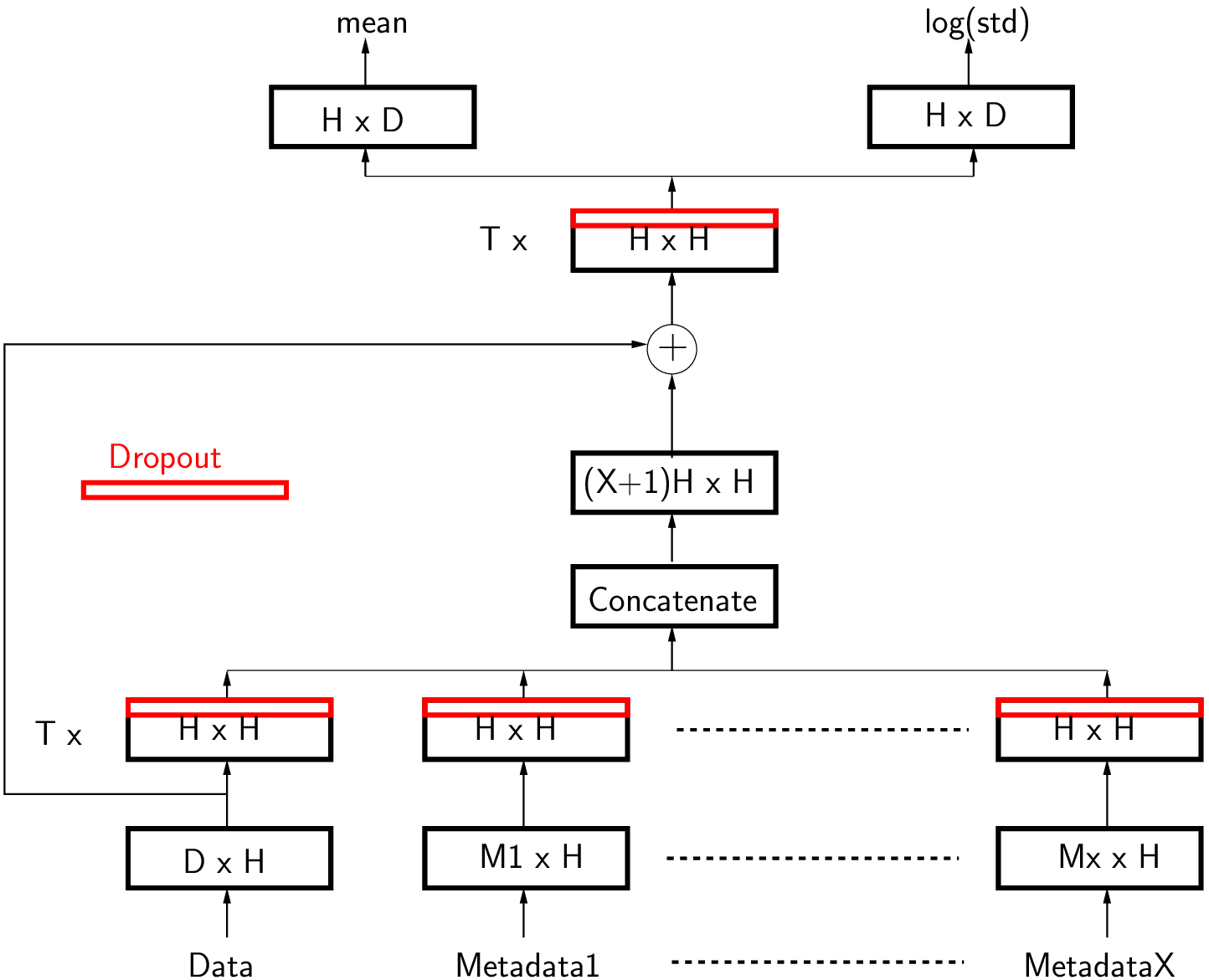,width=8.0cm}\\
    \end{center}
  \end{minipage}
  \caption{A schematic of the DNN used in both examples discussed in section \ref{sec:results}. The exact parameters for $\sf D$, $\sf H$, $\sf Mx$, and $\sf T$ vary for each example.\label{fig:dnn}}
\end{figure}

In Fig. \ref{fig:dnn} we provide a rough schematic of the DNN used in the examples in section \ref{sec:results}. All layers in the DNN use dense linear layers with sigmoid linear unit (SiLU) activation (but no activation in the last layer). The dimensions of each linear layer in Fig. \ref{fig:dnn} are shown. We use $\sf D$ to denote the input dimension, $\sf H$ to denote the hidden dimension and $\sf Mx$ to denote the metadata dimension, respectively. If the metadata dimension is higher than $H$, $H$ is replaced by the metadata dimension for that particular flow (until concatenation). Depending on the example, there may be different number of metadata types ($2$ for the linear model and $4$ for the radio interferometric model). We can also change the depth of the DNN in Fig. \ref{fig:dnn} by changing the parameter $\sf T$ (indicating $\sf T$ linear blocks are appended sequentially, with SiLU activation). Optionally, we introduce dropout at two locations in the DNN (with dropout rate set at 0.1). A residual connection is also created between the input data and the output after concatenation.

We use the AdamW optimizer \cite{AdamW} for training the DNN models. We use learning rate scheduling with a warmup period and cosine decay thereafter. Additional details of the training are given in Table \ref{tab:training}.

\begin{table}[b]
\caption{\label{tab:training}%
DNN and training parameters
}
\begin{ruledtabular}
\begin{tabular}{lcc}
&
Linear model&
Radio interferometric\\
& & model\\
\colrule
Depth $\sf T$& 5 & 1 \\
Hidden dimension $\sf H$ & 256 & 5500 \\
Dropout & no & yes \\
Data normalization & no & yes \\
Epochs & 100 & 200 \\
Batch size & 256 & 256\\
Learning rate & 0.0001 & 0.0001\\
Warmup iterations & 5000 & 5000\\
  Cadence $C$ & 100 & 100 \\
  $\gamma$ & 1.1 & 0.9\\
  $\rho$ & 50 & 1\\
  Pre training & no & yes \\
\end{tabular}
\end{ruledtabular}
\end{table}

\bibliography{references}

\end{document}